\documentclass[journal]{IEEEtran} 

\ifCLASSINFOpdf
\else
\fi
\usepackage{cite}
\usepackage{algorithm,algorithmicx}
\usepackage{algpseudocode}
\usepackage[small,bf]{caption}
\usepackage[cmex10]{amsmath}
\usepackage{amssymb,latexsym,epsfig,subfigure,epic,amscd,mathrsfs,euscript,eufrak,bm}
\usepackage{color}
\usepackage{booktabs}
\usepackage{array}

\usepackage{amsfonts}
\usepackage{amsopn}
\usepackage{graphicx}
\usepackage{url} 
\usepackage{empheq}
\usepackage{epstopdf}

\newtheorem{myprop}{\bf{Proposition}}

\newtheorem{remark}{\bf{Remark}}
\newtheorem{lemma}{\bf{Lemma}}
\newtheorem{corollary}{\bf{Corollary}}
\newcommand{\argmin}{\operatornamewithlimits{arg\,min}}
\newcommand{\argmax}{\operatornamewithlimits{arg\,max}}

 \newcommand{\RN}[1]{%
  \textit{{\uppercase\expandafter{\romannumeral#1}}}%
}

\DeclareMathOperator*{\minimize}{\text{minimize}}

\DeclareMathOperator*{\st}{\text{subject to}}
\DeclareMathAlphabet\mathbfcal{OMS}{cmsy}{b}{n}
\newcommand{\Def}[0]{\mathrel{\mathop:}=}

\IEEEoverridecommandlockouts

\begin{document}
%
\title{{{Accelerated Distributed Dual Averaging over    Evolving Networks of Growing Connectivity
}}}

%

\author{Sijia~Liu,~\IEEEmembership{Member,~IEEE,}
        Pin-Yu~Chen,~\IEEEmembership{Member,~IEEE,}
        and~Alfred~O.~Hero,~\IEEEmembership{Fellow,~IEEE}
\thanks{Copyright (c) 2015 IEEE. Personal use of this material is permitted. However, permission to use this material for any other purposes must be obtained from the IEEE by sending a request to pubs-permissions@ieee.org.}   
\thanks{S. Liu and A. O. Hero are with the Department of Electrical Engineering and Computer Science, University of Michigan, Ann Arbor, E-mail: \{lsjxjtu, hero\}@umich.edu.  P.-Y. Chen is with IBM Thomas J. Watson Research Center, Yorktown Heights, Email: pin-yu.chen@ibm.com }
\thanks{This work was partially supported by grants from the US Army Research Office, grant numbers W911NF-15-1-0479 and W911NF-15-1-0241. Portions of this work were presented at the $42$nd IEEE International Conference on Acoustics, Speech and Signal Processing, New Orleans, USA, in  2017.
}
}


\maketitle

\begin{abstract}
We consider the problem of accelerating distributed optimization in multi-agent networks  by sequentially adding edges. 
Specifically, we extend 
the distributed dual averaging (DDA) subgradient algorithm to evolving networks of growing connectivity and analyze the corresponding improvement in  convergence rate.
 It is known that the convergence rate of DDA is influenced by  the algebraic connectivity of the underlying  network, where better connectivity leads to faster convergence.  However, the impact of network topology design on the convergence rate of DDA has not been fully understood.
In this paper,  we begin by designing  network topologies   via edge selection and scheduling \textcolor{black}{in an offline manner}. 
For edge selection, we determine the best set of candidate edges that achieves the optimal tradeoff between the growth of network connectivity and the usage of network resources. The dynamics of network evolution is then incurred by   edge scheduling. 
Furthermore, we   provide  a tractable approach to analyze the improvement in the convergence rate of DDA induced by the growth of  network connectivity.  Our analysis 
reveals the connection between network topology design and the convergence rate of DDA, {and provides quantitative evaluation of  DDA acceleration for distributed optimization that  is {absent} in the existing  analysis.}
Lastly, numerical experiments show that   DDA can be significantly accelerated using a sequence of well-designed networks, and our  theoretical predictions are well matched to its empirical convergence behavior. 
\end{abstract}


%
\IEEEpeerreviewmaketitle

\section{Introduction}
\label{sec:intro}

{\color{black}
Optimization over networks
 has found a wide range of applications in parallel computing \cite{hendrickson2000graph,pothen1997graph}, sensor networks \cite{blatt2007convergent,olffaxmur07},
communication systems \cite{Kelly1998,chilow07},  machine learning \cite{man93}
 and power grids \cite{dorchebul13}. 
 Many such problems {fall under the framework of} distributed optimization in order to minimize the network loss function, given as   a sum of local objective functions accessed by agents, e.g., cores in massive parallel computing.   The  ultimate goal  is to achieve the solution of  the network-structured optimization problem by using only local information  and in-network communications.  

 In  this paper, we investigate the impact of network topology design on the convergence rate of distributed optimization algorithms. To be specific, 
we focus  on  the convergence analysis of the distributed
dual averaging (DDA) subgradient algorithm  \cite{ducaga12} under \textit{well-designed}
time-evolving networks,  where edges are sequentially added.
\textcolor{black}{There are several motivations for DDA. First, DDA provides a   general distributed optimization framework that can handle problems involving nonsmooth   objective functions and constraints. In contrast, some recently developed accelerated distributed   algorithms  \cite{jakxavmou14,jakxavmou14TSP, shi2015extra,qu2017harnessing,shi2015proximal}  are limited to  problems satisfying conditions  such as strong convexity,  smoothness, and smooth+nonsmooth composite objectives. 
Second, it was   shown by Nesterov  \cite{nes09}   that 
the use of dual averaging  guarantees  boundedness of the sequence of primal  solution points even in the case of an unbounded feasible set. \textcolor{black}{This is in contrast with  some recent algorithms  \cite{hosseini2014online,suzuki2013dual}, which require   boundedness of  the feasible set  for the convergence analysis.}
Third, \textcolor{black}{
the convergence rate of DDA     \cite{ducaga12} can be
decomposed into two additive terms: one reflecting the suboptimality of first-order algorithms and  the other representing the network effect. The latter has
a   tight dependence on
 the algebraic connectivity \cite{chung96} of the network topology.}
 This enables us to accelerate   DDA  
by  making the network fast mixing
 with increasing network connectivity.}
 However, the use of desired time-evolving networks introduces new challenges in distributed optimization: 
a) How to sequentially add edges to a network to maximize   {DDA acceleration}?  b) How to evaluate  the improvement in
the convergence rate of DDA induced by the growth of network connectivity? 
c) How to {connect the dots}  between the strategy of     topology design and the convergence behavior of DDA?
In this paper, we aim to answer these questions in the affirmative.

}

\subsection{Prior and related work}

There has been a significant amount of research on developing  distributed   computation and optimization algorithms  in multi-agent networked  systems 
\cite{olfmur04,hatmes05,xiaoboyd03,karmoura09,nedolsozd09,tusayd12,kopsadrib16,tsiberath86,nedozd09,lobozd11,sunnedvee10,jakxavmou14,jakxavmou14TSP,shi2015extra,qu2017harnessing,shi2015proximal,jakovetic2016distributed,moklingrib17,weiozd13,aybat2017distributed}.
Early research efforts focus on the design of  consensus algorithms for distributed averaging,   estimation and filtering, where `consensus\rq{} means to reach agreement on a common decision for a group of agents \cite{olfmur04,hatmes05,xiaoboyd03,karmoura09,nedolsozd09,tusayd12}. In \cite{kopsadrib16}, a strict  consensus protocol was relaxed to a
protocol that only ensures
neighborhood 
similarity between agents' decisions. 
Although  the existing distributed  algorithms are essentially different,  they {share} a similar consensus-like step  for information averaging, whose speed is associated with  the spectral properties of the underlying networks.

In addition to distributed averaging, estimation and filtering, many  distributed     algorithms have been developed to  solve 
general  optimization problems
\cite{tsiberath86,nedozd09,lobozd11,sunnedvee10,jakxavmou14,jakxavmou14TSP,shi2015extra,qu2017harnessing,shi2015proximal,jakovetic2016distributed,moklingrib17,weiozd13,aybat2017distributed}.
In \cite{tsiberath86,nedozd09,lobozd11,sunnedvee10},   distributed subgradient algorithms involving the consensus step 
were studied  under various scenarios, e.g., networks with  communication delays, dynamic topologies, link failures, and  push-sum averaging protocols. 
In \cite{jakxavmou14} and \cite{jakxavmou14TSP}, fast distributed gradient methods were proposed under both deterministically and randomly time-varying networks. It was shown that
  a faster convergence rate can be achieved by  applying
Nesterov\rq{}s acceleration step \cite{Nesterov83} when the objective function is smooth. 
\textcolor{black}{In \cite{shi2015extra} and \cite{qu2017harnessing}, accelerated first-order   algorithms were proposed, which  utilize the historical information and harness  the function smoothness to obtain a better   estimation of  the gradient. In \cite{shi2015proximal}, a distributed proximal gradient algorithm    was   developed for smooth+nonsmooth composite optimization. However,   
the complexity of this algorithm  is determined by the complexity of implementing the proximal operation, which could be computationally expensive for some forms of the    nonsmooth term in the objective function. 
In \cite{jakovetic2016distributed}, under the assumption of a static network, a probabilistic   node selection scheme was proposed to accelerate the   distributed subgradient algorithm in terms of reducing
  the overall per-node communication  and  gradient computation cost. 
Different from  \cite{jakxavmou14,jakxavmou14TSP, shi2015extra,qu2017harnessing,shi2015proximal,jakovetic2016distributed}, we consider a more general problem setting 
that allows both the 
objective functions and constraints  to be non-smooth.
 In addition, we perform distributed optimization  over a dynamic changing  network topology, and we
consider the resource limited scenario where good network topology design is used to optimize performance (minimize error) subject to constraints on the number of communication links in the network.  We reveal  how the edge addition  scheme affects the network mixing time  and  the  error of a distributed optimization algorithm.
}
  In addition to the first-order   algorithms, distributed implementations   of   Newton method  and an operator splitting method,    alternating direction method of multipliers (ADMM), 
were developed  in 
 \cite{moklingrib17,weiozd13,aybat2017distributed}. Compared to  the first-order algorithms,  second-order methods and ADMM  have faster convergence rate but   at the cost of higher computational complexity.

\textcolor{black}{
  DDA \cite{ducaga12}  has   attracted recent attention  in signal processing, distributed control,   and online  learning  \cite{tsirab12,tsilawrab12,YuaXuZha12,leenedrag16,HosChaMes16}.
}
In \cite{tsirab12}, it was shown that
DDA is robust to communication delays, and the resulting  error decays at a sublinear rate.
In \cite{tsilawrab12}, a push-sum consensus step was combined with DDA that leads to several appealing convergence  properties, e.g., the sublinear convergence rate without the need of a doubly stochastic protocol. 
In \cite{YuaXuZha12}, DDA was studied when the communications among agents are either deterministically or probabilistically quantized.
In \cite{leenedrag16,HosChaMes16}, online versions of DDA were developed when    uncertainties exist in the environment so that  the   cost function is  unknown to local agents prior to decision making. Reference \cite{leenedrag16}  utilized coordinate updating protocols in DDA 
when the global objective function is inseparable.   In  \cite{HosChaMes16},   a distributed weighted dual averaging algorithm was proposed for online learning under switching networks. 


\subsection{Motivation and contributions}

Since the convergence rate of DDA relies on the algebraic connectivity  of   network topology \cite{ducaga12},   it is possible to accelerate   DDA by using well-designed networks. 
In this paper, the desired networks are assumed to possess  growing algebraic connectivity in time. 
The rationale behind that is elaborated on as below.
First, it was {empirically observed } in our earlier work \cite{liucheher17}  that the growth of network connectivity could significantly improve
 the convergence rate of DDA   compared to the case of using a static   network. Therefore,  
distributed optimization can {initially} be performed over a
sparse network of low computation and communication cost, and then  {be accelerated} via a sequence
of well-designed networks for improved convergence. {This is particularly appealing to applications in distributed estimation/optimization with convergence time constraints.}  
Second,
the case of growing connectivity is inspired by  real-world scenarios.
One compelling example is in adaptive mesh parallel scientific computing where the network corresponds to grid points in the mesh and optimization is performed by solving partial differential equations over a domain that is successively and adaptively refined over the network as time progresses \cite{smith2004domain}.  As another example, 
the accumulated  connectivity of online social networks increases over time when users establish new connections (e.g., time-evolving friendship in Facebook or LinkedIn). Another example arises in the design of resilient hardened physical networks \cite{xuguhi14_cf,CPY14ComMag}, where adding edges or rewiring existing edges increases network connectivity  and robustness to node or edge failures.  

Our main {contributions}   are twofold. First, \textcolor{black}{we propose an offline optimization approach   for topology design via  edge selection and   scheduling.} The proposed approach  is motivated by the fact that the algebraic connectivity   is monotone increasing in the edge set \cite{ghoboy06,boyd06}, and thus a well-connected network can be obtained via edge additions. 
In the stage of edge selection, we determine the best candidate edges that achieve  optimal tradeoffs between the growth of network connectivity and the usage of network resources.
\textcolor{black}{The selected edges \textcolor{black}{determined by our offline optimization method} are then scheduled   to update the   network topology for improved convergence of DDA. 
Second,
we provide a general convergence analysis  that reveals the connection between the solution of  topology design  and the convergence rate of DDA.
This leads to a tractable approach to analyze the improvement in the convergence rate of DDA induced by the growth of
network connectivity.}
Extensive numerical results show that 
our theoretical predictions are well matched to   the empirical convergence behavior of the accelerated  DDA algorithm under well-designed networks.

\subsection{Outline}
The rest of the paper is organized as follows. In Section\,\ref{sec: background}, we introduce the concept of distributed optimization and DDA. In Section\,\ref{sec: prob},
we formulate two types of problems for edge selection, and provide a dynamic model to incorporate   the effect of edge scheduling on  DDA. In
Section\,\ref{sec: edge_sel}, we develop efficient optimization methods to solve the problems of edge selection. In Section\,\ref{sec: conv_DDA}, we analyze the convergence properties of DDA under evolving networks of growing connectivity.
In Section\,\ref{sec: numerical}, 
 we demonstrate the effectiveness of DDA over well-designed networks through numerical examples. Finally, in
Section\,\ref{sec: conclusion} we summarize our work and discuss future research
directions.

\section{{Preliminaries}: Graph, Distributed Optimization and Distributed Dual Averaging}\label{sec: background}
 In this section, we provide a background  on  the graph representation of a multi-agent network, the distributed optimization problem, and the distributed dual averaging algorithm.

\subsection{Graph representation}
A graph yields a succinct  representation of  interactions among agents or sensors over a network. 
Let $\mathcal G_t = (\mathcal V, \mathcal E_t)$ denote a \textit{time-varying} undirected unweighted graph at time $t = 0,1,\ldots,T$, where $\mathcal G_0$ is the initial graph, $\mathcal V$ is a node set with cardinality $|\mathcal V| = n$,   $\mathcal E_t \subseteq [n] \times [n] $ is an edge set with cardinality $|\mathcal E_t| = m_t$,  
and $T$ is the length of time horizon. Here for  ease of notation, we  denote by $[n]$ the integer set $\{ 1,2,\ldots, n\}$. 
Let $l \sim (i,j)$ denote the $l$th edge 
of $\mathcal E_t$ that connects  nodes $i$ and   $j$. Without loss of generality,  the ordering of edges is  given  \textit{a priori}. The set of neighbors of node $i$ at time $t$ is denoted by   $\mathcal N_i(t) = \{ j ~|~ (i,j) \in \mathcal E_t \} \cup \{ i \}$. 

A graph is commonly  represented   by an adjacency matrix or a graph Laplacian matrix. Let $\mathbf A_t$ be 
the adjacency matrix of $\mathcal G_t$, where $[\mathbf A_t]_{ij} = 1$ for $(i,j) \in \mathcal E_t$ and $[\mathbf A_t ]_{ij} = 0$, otherwise. Here $[\mathbf X]_{ij}$ (or $\mathbf X_{ij}$) denotes the $(i,j)$-th entry of a matrix $\mathbf X$.
 The graph Laplacian matrix is defined as $\mathbf L_t = \mathbf D_t - \mathbf A_t$, where $\mathbf D_t$ is a degree diagonal matrix, whose $i$-th diagonal entry is given by $\sum_{j} [\mathbf A_t]_{ij}$. 
The graph Laplacian matrix is  positive semidefinite, and has a zero eigenvalue $\lambda_{n} = 0$   with eigenvector $\textcolor{black}{(1/\sqrt{n})}\mathbf 1$. \textcolor{black}{Here $\mathbf 1$ is the column vector of ones.}
In this paper, eigenvalues are sorted in \textit{decreasing} order of magnitude, namely, $\lambda_i \geq \lambda_j$ if $i \leq j$.

The second-smallest  eigenvalue $\lambda_{n-1}(\mathbf L_t)$ of $\mathbf L_t$  is known as the algebraic connectivity   \cite{chung96}, which is positive if and only if the graph  is connected, namely,  there exists a  path between every pair of distinct nodes.  
 In this paper we assume that  $\mathcal G_t$ is connected at each time. 
\textcolor{black}{As will be evident later, it is beneficial to express the graph Laplacian matrix as a product of incidence matrix
\begin{align}
\mathbf L = \mathbf H \mathbf H^T = \sum_{l=1}^{m} \mathbf a_l \mathbf a_l^T, \label{eq: L_inc}
\end{align} 
where for notational simplicity, we drop the time index without loss of generality, 
$\mathbf H = [\mathbf a_1, \ldots, \mathbf a_{m}] \in \mathbb R^{n \times m}$ is the incidence matrix of graph $\mathcal G$ with $n$ nodes and $m$ edges,  and
$\mathbf a_l$ is known as an edge vector corresponding to  the edge $l \sim (i,j)$ with entries $[ \mathbf a_l ]_i = 1$, $[ \mathbf a_l ]_j = -1$ and $0$s elsewhere.
}


\subsection{Distributed optimization}
We consider a convex optimization problem, where the   network cost, given by a sum of local objective functions at agents/sensors, is minimized subject to a convex constraint. That is,
\begin{align}
\begin{array}{ll}
\displaystyle \minimize & \displaystyle f(\mathbf x) \Def \frac{1}{n} \sum_{i=1}^n f_i(\mathbf x) \\
\st & \mathbf x \in \mathcal X,
\end{array}
\label{eq: prob_dist}
\end{align}   
where $\mathbf x \in \mathbb R^p$ is the optimization variable, $f_i$ is convex and $L$-Lipschitz continuous\footnote{The $L$-Lipschitz continuity of $f$ with respect to   a generic norm $\| \cdot \|$ is defined by $| f(\mathbf x) - f(\mathbf y)| \leq L \| \mathbf x - \mathbf y\|$, for $\mathbf x, \mathbf y \in \mathcal X$.},
and $\mathcal X$ is a closed ({not necessarily
bounded}) convex set containing the origin. 
A concrete example of \eqref{eq: prob_dist}   is a distributed estimation problem, where $f_i$ is a square loss and $\mathbf x$ is an unknown parameter to be estimated.
The graph $\mathcal G_t$ imposes communication constraints on distributed optimization to solve problem \eqref{eq: prob_dist}: Each node $i$ only accesses to the local cost function $f_i$ and can communicate directly only with nodes in its neighborhood $\mathcal N_i(t)$ at time $t$.

\subsection{Distributed dual averaging (DDA)}

Throughout this paper we employ the DDA algorithm, which was first proposed in  \cite{ducaga12}, to solve problem \eqref{eq: prob_dist}. 
In DDA, each node $i \in \mathcal V$ iteratively performs the following two steps  
\begin{align}
&\mathbf z_i (t+1) = \sum_{j \in \mathcal N_i(t)} [\mathbf P_t]_{ji} \, \mathbf z_j(t)+ \mathbf g_i(t) 
\label{eq: average_step}\\
&\mathbf x_i(t+1) 
= \argmin_{\mathbf x \in \mathcal X} \left \{ 
 \mathbf z_i (t+1)^T \mathbf x + \frac{1}{\alpha_t} \psi(\mathbf x)
\right \}, \label{eq: update_step}
\end{align}
where $\mathbf z_i(t) \in \mathbb R^d$ is a newly introduced variable for node $i$ at time $t$, $\mathbf P_t \in  \mathbb R^{n \times n}$ is a matrix of non-negative weights that preserves the zero structure of the graph Laplacian $\mathbf L_t$, 
$\mathbf g_i(t)$ is a subgradient of $f_i(\mathbf x)$ at $\mathbf x = \mathbf x(t)$,
$\psi(\mathbf x)$ is a regularizer for stabilizing the update, and $\{ \alpha_t\}_{t=0}^{\infty}$ is a non-increasing sequence of positive step-sizes. 

In  \eqref{eq: update_step}, $\psi(\mathbf x)$ is also known as a proximal function, which is assumed to be $1$-strongly convex with respect to  a generic norm $\| \cdot \|$, and $\psi(\mathbf x) > 0$ and $\psi(\mathbf 0) = 0$. In particular, when $\| \cdot \|$ is the   $\ell_2$ norm, we obtain the canonical proximal function $ \psi(\mathbf x) = (1/2) \| \mathbf x\|_2^2$.
The weight matrix $\mathbf P_t$  in  \eqref{eq: average_step} is   assumed to be doubly stochastic, namely, $\mathbf 1^T \mathbf P_t = \mathbf 1^T$ and $ \mathbf P_t \mathbf 1= \mathbf 1$. A common choice of $\mathbf P_t$ that respects the graph structure of $\mathcal G_t$ is given by   \cite{ducaga12}
\begin{align}
\mathbf P_t = \mathbf I - \frac{1}{2(1 + \delta_{\max})} 
\mathbf L_t ,\label{eq: P_t}
\end{align} 
where 
$\delta_{\max}$ is the maximum degree of   $\{ \mathcal G_t \}$, and $\mathbf P_t$ is  positive semidefinite \cite{LevinPeresWilmer2006}. 
It is worth mentioning that if $\mathbf L_t = \mathbf L_0$ for $t \in [T]$, DDA is performed under a static network \cite{tsirab12,tsilawrab12,ducaga12}. 



\section{Problem Formulation}\label{sec: prob}
In this section, we  begin by  introducing the concept  of growing connectivity. We then formulate the problem of   edge selection, which strikes 
a balance between   network connectivity and network resources. 
Lastly, we  link DDA with the procedure of  edge scheduling. 

%
%

\subsection{Evolving networks of growing connectivity}

 It is known  from \cite{Fiedler1973} that the algebraic connectivity  provides   a lower bound on the node/edge connectivity, which is the least number of node/edge removals that disconnects the graph. 
We refer to $  \mathcal G_t $  as  a dynamic  network  of growing connectivity if  its   algebraic connectivity   is monotonically increasing in time, namely,
\begin{align}
0 < \lambda_{n-1}(\mathbf L_0) \leq \lambda_{n-1}(\mathbf L_1) \leq \ldots \leq \lambda_{n-1}(\mathbf L_T). \label{eq: lam_time_L}
\end{align}
Based on \eqref{eq: P_t}, we equivalently obtain
\begin{align}
\sigma_{2}(\mathbf P_0) \geq \sigma_{2}(\mathbf P_1) \geq \ldots \geq \sigma_{2}(\mathbf P_T), \label{eq: P_inc}
\end{align}
where $\sigma_{2}(\mathbf P_t)$ is the second-largest singular value of $\mathbf P_t$. 
\textcolor{black}{In this paper,
we aim to improve the convergence rate of DDA with the aid of designing  networks of growing connectivity. 
}

\subsection{Edge selection}
Since the algebraic
connectivity   is monotone increasing in the edge set, a well-connected network can be obtained via edge additions \cite{ghoboy06}. 
Given an initial graph $\mathcal G_0$, it is known from \eqref{eq: L_inc} that the Laplacian matrix of the modified graph by adding new edges can be expressed as 
\begin{align}
\mathbf L (\mathbf w) = \mathbf L_0 + \sum_{l=1}^{K} w_l \mathbf a_l \mathbf a_l^T,   ~ \mathbf w \in \{ 0,1\}^K, \label{eq: L_new}
\end{align}
where
\textcolor{black}{$K = n(n-1)/2 - m_0$ is the maximum number of   edges that can be added to $\mathcal G_0$, namely, 
edges in the complement  of $\mathcal G_0$,
  $w_l \in \{ 0, 1 \}$ is a Boolean   variable  indicating  whether or not the $l$th candidate edge is selected,  and $\mathbf a_l \in \mathbb R^n $ is  the  edge vector  corresponding to the added edge $l$ defined in \eqref{eq: L_inc}.
}


Edge addition makes a network  better-connected but consumes more network resources, e.g., bandwidth and energy for inter-node communications.
Therefore, we formulate an   optimization problem  for edge selection, which 
 seeks the optimal tradeoff between  the network connectivity and the resource consumption due to edge addition. That is,
\begin{align}
\begin{array}{ll}
\displaystyle \minimize_{\mathbf w \in \mathbb R^K} & n - \lambda_{n-1}\left (\mathbf L (\mathbf w)   \right ) + \gamma \mathbf c^T \mathbf w \\
\st &  \mathbf w \in \{ 0,1\}^K,
\end{array}
\label{eq: opt_connectivity}
\end{align} 
where $\mathbf L (\mathbf w) $ is given by \eqref{eq: L_new}, 
the term $n - \lambda_{n-1}  (\mathbf L (\mathbf w)    ) $ represents the \lq{}distance\rq{}  to the maximum algebraic connectivity $n$
determined by the complete graph, $\mathbf c = [c_1 ,\ldots, c_K]^T$, $c_l$  is a known   cost associated with the $l$th edge, e.g.,   communication cost   proportional to the distance of  two nodes,  and $\gamma > 0$ is a   regularization parameter  to characterize 
the relative importance of achieving a large network connectivity
versus consuming a few    network resources. Note that if $\mathbf c = \mathbf 1$ in \eqref{eq: opt_connectivity}, the solution of the problem yields a tradeoff between the network connectivity and  the number of selected edges. 

If the number of selected edges (denoted by $k$) is given \textit{a priori}, we can modify problem \eqref{eq: opt_connectivity}  as  
\begin{align}
\begin{array}{ll}
\displaystyle \minimize_{\mathbf w \in \mathbb R^K} & n- \lambda_{n-1}(\mathbf L(\mathbf w) ) + \gamma \mathbf c^T \mathbf w \\
\st &  \mathbf w \in \{ 0,1\}^K,\quad \mathbf 1^T \mathbf w = k.
\end{array}
\label{eq: opt_connectivity_variant}
\end{align}
Different from \eqref{eq: opt_connectivity},     
by varying the regularization parameter $\gamma$, the solution of problem \eqref{eq: opt_connectivity_variant}   strikes a balance between the  network connectivity and  the  communication cost due to edge \textit{rewiring} rather than edge \textit{addition} under the candidate edge set. 

\textcolor{black}{
We remark that problems \eqref{eq: opt_connectivity}-\eqref{eq: opt_connectivity_variant} can be solved offline prior to distributed optimization
using edges in the set complentary to   the initial graph $\mathcal G_0$, the regularization parameter $\gamma$, the edge cost $\mathbf c$, and/or the number of selected edges $k$. 
The formulation \eqref{eq: opt_connectivity}-\eqref{eq: opt_connectivity_variant} specifies the best locations to add the edges, since   the optimization variable $\mathbf w$ encodes which candidate edge is selected.
In \eqref{eq: opt_connectivity}-\eqref{eq: opt_connectivity_variant}, 
the regularization parameter $\gamma$ controls the penalty on communication cost. The regularization term could be replaced with a `hard' constraint on the communication cost, $\mathbf c^T \mathbf w \leq \eta$, where $\eta$ is the maximum cost to be consumed, and similar to   $\gamma$,  a  tradeoff  between   network connectivity and communication cost can be achieved by varying $\eta$. 
}


\subsection{Edge scheduling versus DDA}
\textcolor{black}{Given the solution  of problem \eqref{eq: opt_connectivity} or \eqref{eq: opt_connectivity_variant}, 
we propose an algorithm for scheduling  edge addition,    generating a time-evolving network of growing connectivity.}  
And thus, the problem of \textit{edge scheduling}
arises.
For ease of  updating network,  we assume that  at most one  edge is added at each time step. Such an assumption is commonly used in the design of scheduling protocols for resource management
 \cite{Jiming2011,Gupta04,Gupta06}.
The graph Laplacian  then follows the   dynamical model 
\begin{align}
\mathbf L_t = \mathbf L_{t-1} + u_t \mathbf a_{l_t} \mathbf a_{l_t}^T, ~ u_{t} \in \{ 0 ,1 \},~ l_t \in \mathcal E_{\mathrm{sel}} - \mathcal E_{t-1}
\label{eq: Lt_dyn}
\end{align}
where $u_t$ encodes whether or not a selected edge is added  to $\mathbf L_{t-1}$, $ \mathbf a_{l_t} $  is an edge vector,  and $l_t$ is the edge index among the candidate edges $\mathcal E_{\mathrm{sel}} - \mathcal E_{t-1}$. Here $ \mathcal E_{\mathrm{sel}} \Def \{  l \sim (i,j)  \, | \, w_l^* = 1\}$,   
$\mathbf w^*$ is  the solution  of problem \eqref{eq: opt_connectivity} or \eqref{eq: opt_connectivity_variant},
$\mathcal E_{t-1} $ is the edge set corresponding to $\mathcal G_{t-1}$, and $\mathcal E_{\mathrm{sel}} - \mathcal E_{t-1}$ is 
 the set of all elements that are members of $\mathcal E_{\mathrm{sel}}$ but not members of $\mathcal E_{t-1}$.

\begin{figure}[htb]
\centerline{ \begin{tabular}{c}
\includegraphics[width=.485\textwidth,height=!]{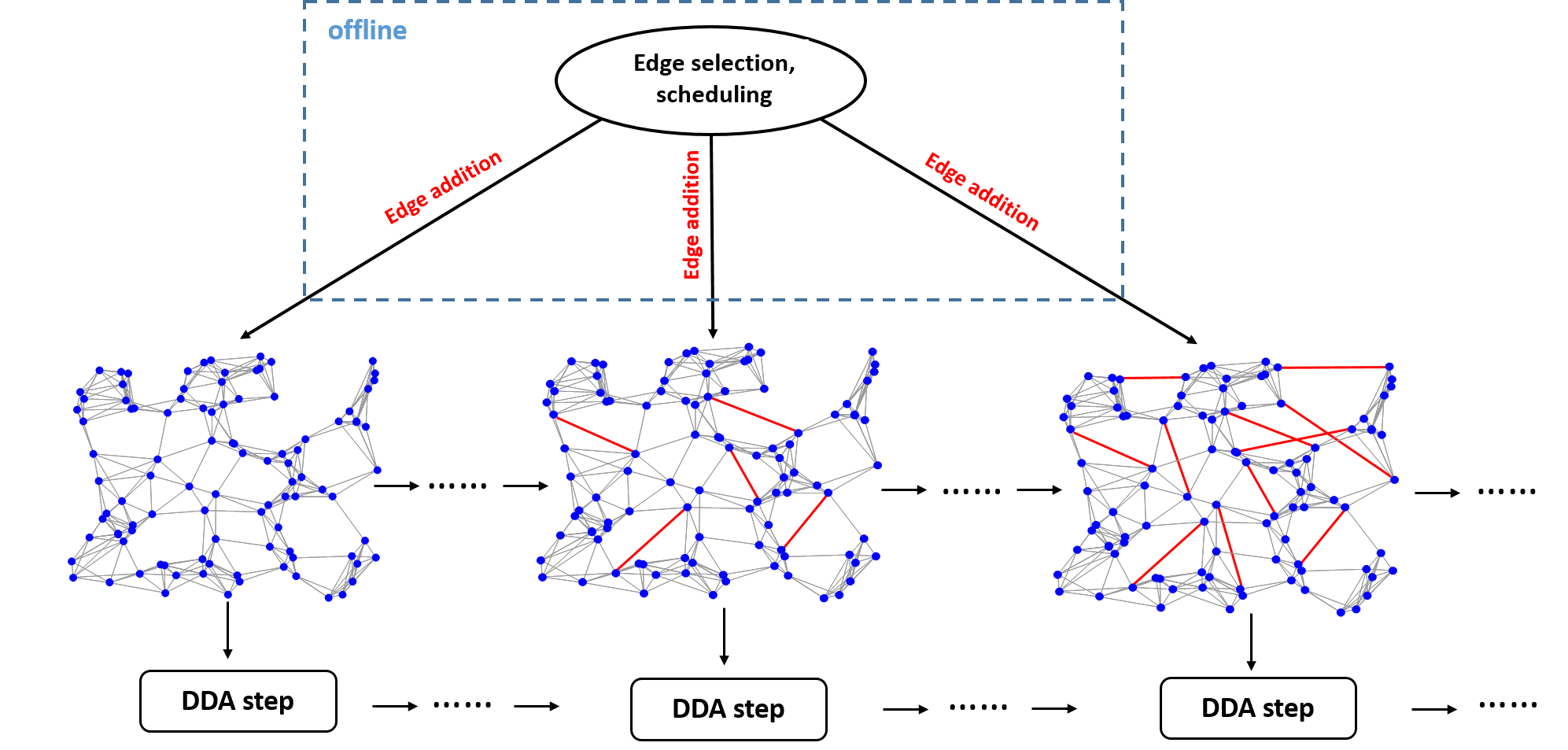}
\end{tabular}}
\caption{\footnotesize{
DDA in conjunction with topology design.
}}
  \label{fig: DDA}
\end{figure}

\textcolor{black}{
Based on  \eqref{eq: Lt_dyn}, we seek the optimal scheduling scheme $\{ u_t \}$ and $\{ l_t \}$ such that the convergence rate of DDA  is the fastest possible. 
Similar to solving problem \eqref{eq: opt_connectivity} or \eqref{eq: opt_connectivity_variant}, an offline   greedy edge scheduling algorithm will be used. 
We summarize the   DDA algorithm in conjunction with network topology design in 
 Fig.\,\ref{fig: DDA}.
}
%
 
\section{Optimization Methods for Topology Design} \label{sec: edge_sel}
In this section, we first show that edge selection problem \eqref{eq: opt_connectivity} and \eqref{eq: opt_connectivity_variant}   can be solved via semidefinite programing.  To improve the computational efficiency,  we then present the projected subgradient algorithm and the greedy algorithm that scale more gracefully with problem size.
\textcolor{black}{Lastly, we discuss the decentralized edge selection method.} 


We begin by reformulating    problems \eqref{eq: opt_connectivity} and \eqref{eq: opt_connectivity_variant} as
\begin{align}
\begin{array}{ll}
\displaystyle \minimize_{\mathbf w \in \mathbb R^K} & n\lambda_{1}\left (\mathbf P(\mathbf w) -{\mathbf 1 \mathbf 1^T}/{n} \right )  + \gamma \mathbf c^T \mathbf w\\
\st &  \mathbf w \in \mathcal T_1, ~ \text{or}~~\mathbf w \in \mathcal T_2,
\end{array}
\label{eq: opt_prob_lam1}
\end{align} 
where 
$\mathbf P(\mathbf w) \Def \mathbf I  - (1/n) \mathbf L(\mathbf w)$, which is positive semidefinite since 
$\lambda_1(\mathbf L(\mathbf w)) \leq n$
\cite{zhao11}, and $\mathcal T_1$  and $\mathcal T_2$ denote   constraints  of problems \eqref{eq: opt_connectivity} and \eqref{eq: opt_connectivity_variant}, respectively.
In the objective function of problem \eqref{eq: opt_prob_lam1}, we have used the fact that   
\begin{align}
\lambda_{n-1}(\mathbf L(\mathbf w)) &= n[1 - \lambda_2(\mathbf P(\mathbf w))] \nonumber \\
&=n - n \lambda_1(\mathbf P(\mathbf w)-{\mathbf 1 \mathbf 1^T}/{n} ). \label{eq: LwPw}
\end{align}
The rationale behind introducing $\mathbf P(\mathbf w)$ is that we can express the second-smallest eigenvalue  $\lambda_{n-1}(\mathbf L(\mathbf w)) $ in the form of the maximum eigenvalue $\lambda_1(\mathbf P(\mathbf w)-{\mathbf 1 \mathbf 1^T}/{n} )$.   The latter is more computationally efficient.  

Moreover  if   the Boolean constraint $\mathbf w \in \{ 0,1\}^K$ is relaxed  to its convex hull $\mathbf w \in [0,1]^K$, we  obtain the   convex program  
\begin{align}
\begin{array}{ll}
\displaystyle \minimize_{\mathbf w \in \mathbb R^K} & n\lambda_{1}\left (\mathbf P(\mathbf w) -{\mathbf 1 \mathbf 1^T}/{n} \right )  + \gamma \mathbf c^T \mathbf w\\
\st &  \mathbf w \in \mathcal C_1, ~ \text{or}~~\mathbf w \in \mathcal C_2,
\end{array}
\label{eq: opt_convex}
\end{align} 
where $\mathcal C_1 = \{ \mathbf w \, | \,  \mathbf w \in [ 0,1]^K\}$,  and $\mathcal C_2 = \{ \mathbf w \, |  \,  \mathbf w \in \mathcal C_1, \mathbf 1^T \mathbf w = k\}$.
Problem \eqref{eq: opt_convex} is \textit{convex} since 
$ \lambda_{1}\left ( \mathbf P(\mathbf w) -{\mathbf 1 \mathbf 1^T}/{n} \right ) $ can be described as the pointwise supremum of   linear functions of $\mathbf w$ \cite{boyvan04},
\begin{align}
 \hspace*{-0.1in}\lambda_{1}\left (\mathbf P(\mathbf w) -{\mathbf 1 \mathbf 1^T}/{n} \right ) = \displaystyle \sup_{ \| \mathbf y \|_2 = 1} \{  \mathbf y^T \mathbf P(\mathbf w) \mathbf y - ( \mathbf y^T \mathbf 1 )^2/n  \}. \label{eq: point_linear}
\end{align}
 
\subsection{Semidefinite programing}

In   problem \eqref{eq: opt_convex}, we introduce
 an epigraph variable $s \in \mathbb R$ and rewrite it as
\begin{align}
\begin{array}{ll}
\displaystyle \minimize_{s \in \mathbb R,\mathbf w \in \mathbb R^K} & s  + \gamma \mathbf c^T \mathbf w\\
\st &   n\lambda_{1}\left (\mathbf P(\mathbf w)-{\mathbf 1 \mathbf 1^T}/{n} \right ) \leq s
\vspace*{0.02in} \\
& \mathbf w \in \mathcal C_1, ~ \text{or}~~\mathbf w \in \mathcal C_2,
\end{array}
\label{eq: opt_convex_1}
\end{align} 
where     the first inequality constraint  is satisfied with equality at the optimal   $\mathbf w$ and $s$.
Problem \eqref{eq: opt_convex_1} can be  cast as a semidefinite program (SDP)
\begin{align}
\begin{array}{ll}
\displaystyle \minimize_{s \in \mathbb R,\mathbf w \in \mathbb R^K} & s  + \gamma \mathbf c^T \mathbf w\\
\st &  \mathbf L(\mathbf w)  + (s-n) \mathbf I +   \mathbf 1 \mathbf 1^T \succeq 0  
\vspace*{0.02in} \\
& \mathbf w \in \mathcal C_1, ~ \text{or}~~\mathbf w \in \mathcal C_2,
\end{array}
\label{eq: opt_convex_2}
\end{align} 
where
$\mathbf X \succeq 0$ (or $\mathbf X \preceq 0$) signifies that $\mathbf X$ is positive semidefinite (or negative semidefinite), and
 the first matrix inequality holds due to
\begin{align}
n\lambda_{1}\left (\mathbf P(\mathbf w)-{\mathbf 1 \mathbf 1^T}/{n} \right ) \leq s &\Leftrightarrow \mathbf P(\mathbf w)-{\mathbf 1 \mathbf 1^T}/{n} \preceq s/n \mathbf I \nonumber \\
& \Leftrightarrow n \mathbf I - \mathbf L(\mathbf w) - \mathbf 1 \mathbf 1^T \preceq s\mathbf I. \nonumber
\end{align}
Here we recall from \eqref{eq: opt_prob_lam1} that $\mathbf P(\mathbf w) = \mathbf I  - (1/n) \mathbf L(\mathbf w)$. 

Given the solution of problem \eqref{eq: opt_convex_2}  $\mathbf w^*$, we can then
generate a suboptimal  solution of the original nonconvex problem \eqref{eq: opt_prob_lam1}. If the number of selected edges $k$ is given \textit{a priori} (under $\mathbf w \in \mathcal T_2$), one can choose   edges given by the first $k$ largest
entries of $\mathbf w^*$. Otherwise, we  truncate $\mathbf w^*$ by thresholding   \cite{bacjenmai12}
\begin{align}
\mathbf w = \mathcal I( \mathbf w^* - \rho \mathbf 1 ), \label{eq: w_binary}
\end{align}
where $\rho > 0$ is a  specified threshold, and $\mathcal I(\mathbf x)$ is an indicator function, which is taken elementwise with $\mathcal I(x_i) = 1$ if $x_i \geq 0$, and $0$ otherwise. If $\rho = 0.5$, the operator \eqref{eq: w_binary} signifies the   Euclidean projection of   $\mathbf w^*$ onto the Boolean constraint $\mathcal C_1$.

The interior-point algorithm is commonly used to solving SDPs. 
The computational  complexity 
is approximated by $O(a^2 b^{2.5} + a b^{3.5})$ \cite{nem12}, where $a$ and $b$
denote the number of optimization variables and the size of
the semidefinite matrix, respectively. In \eqref{eq: opt_convex_2}, we have $a = K+1$ and $b = n$, which leads to the complexity $O(K^{2}n^{2.5} + K n^{3.5})$. In the   case of $K  = O(n^2)$, the computational complexity of SDP becomes $O(n^{6.5})$.  
 Clearly, computing solutions to semidefinite programs
becomes inefficient for networks of medium or large size. 

\subsection{Projected subgradient algorithm}
\textcolor{black}{Compared to semidefinite programming, the projected subgradient algorithm   is more computationally attractive 
\cite{boydxiao03report,boyvan04}, which only requires the information about the subgradient of the objective function, denoted by $\phi (\mathbf w)$, in  \eqref{eq: opt_convex}.
Based on \eqref{eq: point_linear}, a subgradient of $\phi(\mathbf w)$ is given by
}
\begin{align}
\frac{\partial \phi(\mathbf w)}{\partial \mathbf w} & = n\frac{\partial (\tilde {\mathbf y}^T  \mathbf P(\mathbf w) \tilde {\mathbf y})}{\partial \mathbf w} + \gamma \mathbf c  = -\frac{\partial ( \tilde {\mathbf y}^T  \mathbf L(\mathbf w) \tilde {\mathbf y} )}{\partial \mathbf w} + \gamma \mathbf c \nonumber \\
& = [-\tilde {\mathbf y}^T \mathbf a_1 \mathbf a_1^T \tilde {\mathbf y}, \ldots, -\tilde {\mathbf y}^T \mathbf a_K \mathbf a_K^T\tilde {\mathbf y} ]^T + \gamma \mathbf c, \label{eq: subgrad}
\end{align}
where 
$\tilde{\mathbf y}$ is the  eigenvector corresponding to the maximum eigenvalue  $ \lambda_{1}\left (\mathbf P(\mathbf w) -{\mathbf 1 \mathbf 1^T}/{n} \right ) $.

The projected subgradient algorithm   iteratively performs 
\begin{align}\hspace*{-0.07in}
\mathbf w^{(\ell+1)} = \left \{
\begin{array}{cc}
\mathcal P_{\mathcal C_1} \left ( \mathbf v
\right ) & \text{for  \eqref{eq: opt_convex} with $\mathcal C_1$}\\
\mathcal P_{\mathcal C_2} \left (
\mathbf v
\right )  & \text{for  \eqref{eq: opt_convex} with $\mathcal C_2$,}
\end{array}
\right.   
\label{eq: proj_grad}
\end{align}
where $\ell$ is the iteration index, 
$\mathbf v = \mathbf w^{(\ell)} - \kappa_\ell
\frac{\partial \phi(\mathbf w^{(\ell)})}{\partial \mathbf w}$,
$\mathbf w^{(0)}$ is a known initial point, $\kappa_\ell$ is the step size (e.g., a diminishing non-summable step size $\kappa_\ell \propto 1/\sqrt{\ell}$), and $\mathcal P_{\mathcal C_i}(\mathbf v)$ represents the Euclidean projection of $\mathbf v$ onto the constraint set $\mathcal C_i$ for $i \in \{ 1,2\}$.

The projection onto
 the box constraint $\mathcal C_1 = \{ \mathbf w \, | \,  \mathbf w \in [ 0,1]^K\}$ yields an elementwise thresholding operator \cite[Chapter\,6.2.4]{parboy13}
 \begin{align}
[ \mathcal P_{\mathcal C_1} (\mathbf v) ]_l = \left \{
\begin{array}{ll}
0 & v_l \leq 0 \\
v_l & v_l \in [0,1], \\
1 & v_l \geq 1
\end{array}
\right.  ~~ l \in [K].
\label{eq: proj_box}
\end{align}
Acquiring $\mathcal P_{\mathcal C_2} (\mathbf v) $ is
  equivalent to finding the minimizer of problem
\begin{align}
\begin{array}{ll}
\displaystyle \minimize_{\mathbf w} & \| \mathbf w - \mathbf v \|_2 \\
\st &  \mathbf w \in [0,1]^K, ~ \mathbf 1^T \mathbf w = k.
\end{array}
\label{eq: proj_variant}
\end{align}
By applying Karush-Kuhn-Tucker (KKT) conditions, the solution of problem \eqref{eq: proj_variant} is  given by  \cite[Chapter\,6.2.5]{parboy13}
 \begin{align}
\mathcal P_{\mathcal C_2} (\mathbf v)  = \mathcal P_{\mathcal C_1} (\mathbf v - \mu \mathbf 1),
\label{eq: proj_box2}
\end{align}
where $\mu$ is the Lagrangian multiplier associated with the equality constraint of  problem \eqref{eq: proj_variant}, which is given by  the root of the equation
\begin{align}
h(\mu)\Def \mathbf 1^T   \mathcal P_{\mathcal C_1} (\mathbf v - \mu \mathbf 1)  - k = 0. \label{eq: mu}
\end{align}
In \eqref{eq: mu}, we note that 
$h(\max(\mathbf v)) = -k \leq 0$ and $h(\min(\mathbf v-\mathbf 1)) = K-k \geq 0$, where $\max(\mathbf x)$  and  $\min(\mathbf x)$ denote  the maximum and minimum entry of $\mathbf x$, respectively. Since $h(\max(\mathbf v))$ and $h(\min(\mathbf v-\mathbf 1)) $ have opposite signs, we can carry out a bisection procedure on $\mu$ to find the
root of \eqref{eq: mu}, denoted by $\mu^*$. The basic idea is that we 
iteratively shrink the lower bound $\check \mu$ and the upper bound   $\hat \mu$ of $\mu^*$ ($\check \mu$ and $\hat \mu$ are initially given by $\min(\mathbf v-\mathbf 1)$ and $\max(\mathbf v)$), so that no point  of $\{ v_l, v_l - 1 \}_{l=1}^K$ lies in $(\check \mu, \hat \mu)$. As a result, we can derive $\mu^*$  from \eqref{eq: mu}    
\begin{align}
\mu^* = \frac{\sum_{l \in \mathcal S} v_l + |\mathcal S_1| - k }{|\mathcal S|},
\end{align}
where 
$
 \mathcal S = \{ l\,|\,v_l- 1 < \check \mu,\,  \hat \mu < v_l\}$  and    $\mathcal S_1 =  \{l\,|\,  \hat \mu < v_l - 1\}$.

The   computational complexity of  the projected subgradient algorithm (per iteration) is dominated by 
the calculation of  the subgradient  in   \eqref{eq: subgrad}, namely, computation of   the maximum eigenvalue  and the corresponding  eigenvector of $\mathbf P(\mathbf w)  -{\mathbf 1 \mathbf 1^T}/{n} $. This leads to the complexity  $O(n^2)$ by using the power iteration method, which can also be performed in a decentralized manner shown in 
 Sec.\,\ref{sec: dec_eig}.
Similar to  \eqref{eq: w_binary}, we can further generate a  Boolean selection vector after applying the projected subgradient algorithm.

\subsection{Greedy algorithm}

If   the number of selected edges  is known \textit{a priori}, \textcolor{black}{the greedy algorithm developed in  \cite{ghoboy06}  can be used to solve problem \eqref{eq: opt_prob_lam1}. 
As suggested in \cite{ghoboy06}, the partial derivative \eqref{eq: subgrad}
  corresponds to 
the predicted decrease in the objective value of \eqref{eq: opt_prob_lam1} due to  adding the $l$th edge. 
We adopt a greedy heuristic to choose one edge  at a time   to maximize the quantity
$
\tilde {\mathbf y}^T \mathbf a_l \mathbf a_l^T \tilde {\mathbf y} -  \gamma c_l
 $ over all remaining  candidate edges. Specifically,
 \begin{align}
l_t^* & = \argmax_{l_t \in \mathcal E_{\mathrm{candidate}} - \mathcal E_{t-1}} ~  
\tilde {\mathbf y}_{t-1}^T  \mathbf a_{l_t} \mathbf a_{l_t}^T\tilde {\mathbf y}_{t-1}
- \gamma c_{l_t} 
\nonumber \\
& =  \argmax_{l_t \in \mathcal E_{\mathrm{candidate}} - \mathcal E_{t-1}} ~  
 {\mathbf v}_{t-1}^T  \mathbf a_{l_t} \mathbf a_{l_t}^T {\mathbf v}_{t-1}
- \gamma c_{l_t} 
, \label{eq: edge_selection_greedy}
\end{align}
for $t = 1,2,\ldots, |\mathcal E_{\mathrm{candidate}} |$,
where $t$ denotes the greedy iteration  index, $\mathcal E_{t-1} $ is the edge set corresponding to $\mathcal G_{t-1}$, and
$\mathcal E_{\mathrm{candidate}}$ denotes the set of candidate edges.} 
In \eqref{eq: edge_selection_greedy},
$\tilde{\mathbf y}_t$ is the  eigenvector corresponding to the maximum eigenvalue  $ \lambda_{1}\left (\mathbf P(\mathbf w_{t-1}) -{\mathbf 1 \mathbf 1^T}/{n} \right ) $, 
 $\mathbf v_{t-1}$  is the eigenvector of $\mathbf L (\mathbf w_{t-1})$ corresponding to  $\lambda_{n-1}(\mathbf L (\mathbf w_{t-1}))$, and
 $\mathbf w_{t-1}$ is the selection vector determined by $\{ l_i^* \}_{i=1}^{t-1}$. The second equality in \eqref{eq: edge_selection_greedy}
 holds due to the relationship between $\mathbf P(\mathbf w_{t-1})$ and $\mathbf L(\mathbf w_{t-1})$
 in \eqref{eq: LwPw}. If $\gamma = 0$,  the greedy update \eqref{eq: edge_selection_greedy} becomes the rule to maximize the algebraic connectivity of a network in \cite{ghoboy06}. 
 At each iteration, the computational complexity of   the greedy algorithm is
the same as the projected subgradient algorithm due to the eigenvalue/eigenvector computation. The overall complexity of the greedy algorithm is given by 
 $O(|\mathcal E_{\mathrm{candidate}} | n^2)$.


{\color{black}{

\subsection{Decentralized implementation}
\label{sec: dec_eig}

Both the projected subgradient algorithm and the greedy algorithm largely rely on the computation of
    the eigenvector $\tilde{\mathbf y}$ associated with the dominant eigenvalue of $\mathbf P(\mathbf w)  -{\mathbf 1 \mathbf 1^T}/{n} $. 
Such a computation can be performed in a decentralized manner based on only local information  $\mathbf w_{\mathcal N_i}$ at each agent $i \in [n]$, namely, the $i$th row of $\mathbf P(\mathbf w) $. Here we recall from \eqref{eq: opt_prob_lam1} that $\mathbf P(\mathbf w) = \mathbf I  - (1/n) \mathbf L(\mathbf w)$. 
In what follows, we will drop  $\mathbf w$ from $\mathbf P(\mathbf w)$ and $\mathbf L(\mathbf w)$ for notational simplicity. 
 Inspired by \cite[Algorithm\,3.2]{xuguhi14_cf}, a decentralized version of power iteration (PI) method,  we propose Algorithm\,1  to compute $\tilde{\mathbf y}$ in a distributed manner.

\begin{algorithm}
\caption{Decentralized computation of $\tilde{\mathbf y}$}
\begin{algorithmic}[1]
\State Input: initial value $\{ \tilde y_{i}(0) \}_{i=1}^n$ for $n$ nodes, row-wise (local) information of $\mathbf P$ at each agent, and
$N_1 , N_2>0$
\For{$s = 0,1,\ldots, N_1$}
\For{$q = 0,1,\ldots, N_2$}
\State \textit{average consensus}: each agent $i \in [n]$  computes
\begin{align}
\boldsymbol \phi_i(q+1) = {\sum_{j \in \mathcal N_i} P_{ij} \boldsymbol \phi_j(q) }, ~\boldsymbol \phi_i(0) =\tilde{ y}_i(s) \mathbf e_i  \label{eq: phi_i}
\end{align}
\EndFor
\State  \textit{PI-like step}: $\boldsymbol \phi_i(s) \leftarrow \boldsymbol \phi_i(q)$, each agent   computes
\begin{align}
\tilde{ y}_i(s+1) = \frac{\sum_{j \in \mathcal N_i} P_{ij} \tilde{ y}_j(s)  - \mathbf 1^T \boldsymbol \phi_i(s) }{n \|\boldsymbol \phi_i(s) \|_2 } \label{eq: yi}
\end{align}
\EndFor  
\end{algorithmic}
\end{algorithm}

The rationale behind Algorithm\,1 is elaborated on as follows. The step \eqref{eq: phi_i} implies that
$
\boldsymbol \phi( q+1) = (\mathbf P \otimes \mathbf I_n) \boldsymbol \phi(  q)
$, where $\boldsymbol \phi(  q) = [ \boldsymbol \phi_1(q)^T, \ldots, \boldsymbol \phi_n(q)^T   ]^T$, and $\otimes$ denotes the Kronecker product. We then have  for each $i \in [n]$,
\begin{align}
\lim_{q\to \infty}\boldsymbol \phi_i(q) & = \lim_{q\to \infty} (\mathbf P^q \otimes \mathbf I_n) \boldsymbol \phi(\mathbf 0) =  ((1/n) \mathbf 1 \mathbf 1^T \otimes \mathbf I_n) \boldsymbol \phi(  0) \nonumber \\
& =  ({1}/{n}) \sum_i \boldsymbol \phi_i(\mathbf 0) =  ({1}/{n})   \tilde{ \mathbf y}(s), \label{eq: ave_consensus}
\end{align}
where $ \tilde{ \mathbf y}(s) = [  \tilde{ y}_1(s), \ldots, \tilde{ y}_n(s)]$, and
we have used  the fact that
$  \lim_{q\to \infty} \mathbf P^q  = \mathbf 1 \mathbf 1^T/n $
according to Perron-Frobenius theorem.
Clearly, the  step \eqref{eq: phi_i} yields the average consensus towards $(1/n)  \tilde{ \mathbf y}(s)$.
Substituting \eqref{eq: ave_consensus} into \eqref{eq: yi}, we have 
\begin{align}
\tilde { \mathbf y} (s+ 1) = \frac{( \mathbf P -  \mathbf 1 \mathbf 1^T/n)  \tilde{ \mathbf y}(s)  }{\| \tilde{ \mathbf y}(s) \|_2}, 
\end{align}
which leads to the power iteration step given the matrix $ \mathbf P -  \mathbf 1 \mathbf 1^T/n$. 

Based on Algorithm\,1, the proposed edge selection algorithms can be implemented in a decentralized manner.  
At the $\ell$th iteration of the projected subgradient algorithm \eqref{eq: proj_grad},  each agent $i \in [n]$  updates
\begin{align}
[\mathbf w]_{ l} ^{(\ell+1)}  = \mathcal P_{\mathcal C_1 } 
\left (
[\mathbf w^{(\ell)}]_{l} -    \kappa_{\ell} \gamma c_l +  \kappa_{\ell}  (\tilde y_i - \tilde y_j)^2   
\right )
\end{align}
for $ l\sim(i,j)\in \tilde{\mathcal E}_i$, where $\tilde{\mathcal E}_i$ is the set of edges connected to agent $i$, and we have used the fact that
$  (\tilde y_i - \tilde y_j)^2  =\tilde {\mathbf y}^T \mathbf a_l \mathbf a_l^T \tilde {\mathbf y} $, and $\mathcal P_{\mathcal C_1 } $ is the elementwise thresholding operator \eqref{eq: proj_box}.
According to \eqref{eq: proj_box2}, 
if we replace the constraint set $\mathcal C_1$ with $\mathcal C_2$, 
an additional step is required, namely, solving the scalar equation \eqref{eq: mu} in a decentralized manner. We then apply the max-consensus protocol, e.g., random-broadcast-max \cite{iutcibjak12,hecheshi14}, to compute 
 $\max( \mathbf v )$ and $\min(\mathbf v-\mathbf 1)$ in \eqref{eq: mu}, where the latter can be achieved via $\max( \mathbf 1 -\mathbf v)$.
After that, a bisection procedure with respect to a scalar $\mu$ could be   performed at   nodes.

Towards the decentralized implementation of the greedy algorithm, it is beneficial to rewrite the greedy 
   heuristic  as
\begin{align}
&\max_{l}  \{  \tilde {\mathbf y}^T \mathbf a_l \mathbf a_l^T \tilde {\mathbf y} - \gamma c_l \} \nonumber \\
= & \max_{i \in [n]} ~ \{  \max_{l \sim (i,j)  \in \tilde{\mathcal E}_i}  \,     (\tilde y_i - \tilde y_j)^2  - \gamma c_l       \}, \label{eq: max_greedy}
\end{align}
where we drop the greedy iteration index for ease of notation.
Clearly, once $\tilde{ \mathbf y}$ is obtained by using Algorithm\,1, each agent can readily compute the inner maximum of \eqref{eq: max_greedy} using local information, and can then
apply a  max-consensus protocol to attain the solution of  \eqref{eq: max_greedy}. 
}}


\section{Convergence Analysis of DDA under Evolving Networks of Growing Connectivity}
\label{sec: conv_DDA}
In this section, we 
show that there exists a tight connection between the
convergence rate of DDA and the edge selection/scheduling scheme. We also   \textcolor{black}{quantify the improvement in convergence of DDA induced by growing connectivity.}




 Prior to studying the impact of  evolving networks  on the convergence  of DDA, 
Proposition\,\ref{prop: inc_connect}
evaluates the increment of network connectivity induced by the edge addition in \eqref{eq: Lt_dyn}. 
\begin{myprop}
\label{prop: inc_connect}
Given the graph Laplacian matrix in \eqref{eq: Lt_dyn}, the increment of network connectivity at consecutive time
steps is lower bounded as
\begin{align}
\lambda_{n-1}(\mathbf L_t) - \lambda_{n-1}( \mathbf L_{t-1} ) \geq
\frac{u_t ( \mathbf a_{l_t}^T \mathbf v_{t-1} )^2}{\frac{6 }{\lambda_{n-2}(\mathbf L_{t-1}) - \lambda_{n-1}(\mathbf L_{t-1})}+ 1}, \label{eq: inc_conn}
\end{align}
where $\mathbf v_{t}$  is the eigenvector of $\mathbf L_t$ corresponding to  $\lambda_{n-1}(\mathbf L_t)$. 
\end{myprop}
\textbf{Proof}: 
See Appendix\,\ref{app: inc_connect}.
\hfill $\blacksquare$ 

\begin{remark}
Reference  \cite{ghoboy06}   also presented  a   lower bound on the algebraic connectivity of a
graph obtained by adding a single edge to a connected
graph. However, our bound in Proposition\,\ref{prop: inc_connect} is
 tighter than that of  \cite{ghoboy06}.
\end{remark}

Proposition\,\ref{prop: inc_connect}  implies that if one edge is scheduled (added) at time $t$,
the least increment of network connectivity can be evaluated by the right hand side of \eqref{eq: inc_conn}. 
\textcolor{black}{This also implies that based on the knowledge at time $t-1$, 
  maximizing $u_t ( \mathbf a_{l_t}^T \mathbf v_{t-1} )^2$ yields an improved network  connectivity at time $t$. Similar to \eqref{eq: edge_selection_greedy}, when $u_t = 1$, a greedy solution of $l_t$ can be obtained by setting 
   $\mathcal E_{\mathrm{candidate}} = \mathcal E_{\mathrm{sel}}$ and $\gamma = 0 $,  
  \begin{align}
l_t^* = \argmax_{l_t \in \mathcal E_{\mathrm{sel}} - \mathcal E_{t-1}} ~  ( \mathbf a_{l_t}^T \mathbf v_{t-1} )^2, \label{eq: edge_schedule}
\end{align}
where  $\mathcal E_{\mathrm{sel}}$ is the set of selected edges given by the solution  of problem \eqref{eq: opt_connectivity} or \eqref{eq: opt_connectivity_variant}. 
The edge  scheduling strategy \eqref{eq: edge_schedule}   can be determined offline, and
as will be evident later, it is consistent with the maximization of  the convergence speed of DDA.
}

With the aid of Proposition\,\ref{prop: inc_connect},
 we  can establish the theoretical connection between 
the edge addition and  the  convergence rate of DDA.
It is known from \cite{ducaga12} that  for each agent $i \in [n]$, the convergence of  the running local average $\hat {\mathbf x}_i (T) = (1/T) \sum_{t=1}^T \mathbf x_i (t)$ to the solution of problem \eqref{eq: prob_dist}, denoted by $\mathbf x^*$, is governed by two error terms:
a) optimization error  common to subgradient algorithms, and b) network penalty  due to message passing. This basic convergence result is shown in Theorem\,1. 

{\textbf{Theorem\,1 \cite[Theorem\,1]{ducaga12}}}: 
Given the updates \eqref{eq: average_step} and \eqref{eq: update_step}, the difference $f( \hat{\mathbf x}_i(T)) - f(\mathbf x^*)$ for $i \in [n]$ is upper bounded as
$
f( \hat{\mathbf x}_i(T)) - f(\mathbf x^*)\leq \mathrm{OPT} + \mathrm{NET}
$. Here
\begin{align}
& \hspace*{-0.05in} \mathrm{OPT} =  \frac{1}{T \alpha_T} \psi(\mathbf x^*) + \frac{L^2}{2T} \sum_{t=1}^T \alpha_{t-1}  , \label{eq: OPT} \\
& \hspace*{-0.05in}\mathrm{NET} = \hspace*{-0.02in} \sum_{t=1}^T\hspace*{-0.02in}\frac{ L \alpha_t}{T} \hspace*{-0.02in} \left ( \hspace*{-0.02in} \frac{2}{n} \hspace*{-0.02in} \sum_{j=1}^n \hspace*{-0.02in} \|  \bar{\mathbf z}(t) - \mathbf z_j(t) \|_* \hspace*{-0.02in} +\hspace*{-0.02in} \| \bar{\mathbf z}(t) - \mathbf z_i(t)\|_* \right ), \label{eq: NET}
\end{align}
where  $\bar {\mathbf z} (t) = (1/n) \sum_{i=1}^n \mathbf z_i (t)$, and $\| \cdot \|_*$ is the dual norm\footnote{$\|\mathbf v\|_*  \Def \mathrm{sup}_{ \| \mathbf u \| = 1} \mathbf v^T \mathbf u$.} to $\| \cdot \|$.  
\hfill $\blacksquare$

In Theorem\,1,  the optimization error   \eqref{eq: OPT}   can be made arbitrarily small for   a sufficiently large time horizon $T$ and a proper step size $\alpha_t$, e.g., $\alpha_t \propto 1/\sqrt{t}$. The network error  \eqref{eq: NET} measures the deviation of each agent\rq{}s local estimate from the average consensus value.
However, it is far more  trivial to bound the network error \eqref{eq: NET} using spectral properties of dynamic networks in \eqref{eq: Lt_dyn}, and to exhibit 
 the impact of topology design (via edge selection and scheduling) on the 
 convergence rate of DDA.


Based on   \eqref{eq: average_step}, we can express $\mathbf z_i(t)$ as
\begin{align}
\mathbf z_i(t+1) 
  =& \sum_{j=1}^n [ \boldsymbol \Phi(t,0)]_{ji} \mathbf z_j(0) \nonumber \\
&+ \sum_{s = 1}^t \left (
 \sum_{j=1}^n [ \boldsymbol \Phi(t,s)]_{ji} \mathbf g_j(s-1)\right ) + \mathbf g_i(t),
\label{eq: z_it}
\end{align}
where without loss of generality  let $\mathbf z_j(0) = \mathbf 0 $, and 
$\boldsymbol \Phi(t,s)$ denotes the product of time-varying doubly stochastic matrices, namely,
\begin{align}
\boldsymbol \Phi(t,s) = \mathbf P_t \mathbf P_{t-1} \times \cdots \times \mathbf P_s,~ s\leq t.
\end{align}
It is worth mentioning that $\boldsymbol \Phi(t-1,s)$ is a doubly stochastic matrix since $\mathbf P_t $ is doubly stochastic. 
As illustrated in Appendix\,\ref{app: ineq} (Supplementary Material), the network error \eqref{eq: NET} can be explicitly associated with the spectral properties of  $\boldsymbol \Phi(t,s) $. That is,
\begin{align}
\| \bar{\mathbf z}(t) - \mathbf z_i(t) \|_* & \leq    L  \sqrt{n} \sum_{s = 1}^{t-1} \sigma_2(\boldsymbol \Phi(t-1,s) )    + 2 L. \label{eq: dif_Phi_new2}
\end{align}


It is clear from \eqref{eq: dif_Phi_new2} that to bound the network error  \eqref{eq: NET}, 
a careful study on  $ \sigma_2(\boldsymbol \Phi(t-1,s) ) $ is essential. 
In Lemma\,\ref{Lemma: sigma2}, we  relate $\sigma_2(\boldsymbol \Phi(t,s))$ to $\{ \sigma_2(\mathbf P_i)  \}_{i=s}^t$, 
where
the latter associates with the    algebraic connectivity   through \eqref{eq: P_t}.


\begin{lemma}\label{Lemma: sigma2}
Given  $\boldsymbol \Phi(t,s) = \mathbf P_t \mathbf P_{t-1} \times \cdots \times \mathbf P_s$, we obtain
\begin{align}
\sigma_2(\boldsymbol \Phi(t,s)) \leq \prod_{i=s}^t \sigma_2 (\mathbf P_i),  \label{eq: sigma2_Phi}
\end{align}
where $\sigma_2 (\mathbf P)$ is the second largest singular value of a matrix $\mathbf P$. 
\end{lemma}
\textbf{Proof:} See Appendix\,\ref{app: sigma2}. 
\hfill $\blacksquare$

Based on \eqref{eq: P_t} and \eqref{eq: inc_conn}, the sequence 
$\{ \sigma_2(\mathbf P_i)  \}$ has the property
\begin{align}
 \sigma_{2}(  \mathbf P_{t}  ) &\leq  \sigma_{2}(  \mathbf P_{t-1}  ) - u_t  b_{t-1} {(\mathbf a_{l_t}^T \mathbf v_{t-1})^2} \nonumber \\
 &\leq  \sigma_{2}(  \mathbf P_{0}  ) - \sum_{i = 1}^t u_i b_{i-1} (  \mathbf a_{l_i}^T \mathbf v_{i-1}  )^2,
\label{eq: single_sigma2t}
\end{align}
where $b_i^{-1} \Def 2(1+\delta_{\max})+12(1+\delta_{\max})/( \lambda_{n-2}(\mathbf L_{i}) - \lambda_{n-1}(\mathbf L_{i}) ) $. Here we recall that $u_t$ is a binary variable to encode whether or not one edge is scheduled at time $t$, and $l_t$ is the index of the new edge involved in DDA.

It is clear from \eqref{eq: dif_Phi_new2} and \eqref{eq: single_sigma2t} that Proposition\,\ref{prop: inc_connect} and
Lemma\,\ref{Lemma: sigma2}  connect  the convergence rate of DDA with 
the edge selection and scheduling schemes. We bound  the network error \eqref{eq: NET} in Theorem\,2.

\begin{myprop}\label{prop: net}
Given the dynamic network \eqref{eq: Lt_dyn}, the length of time horizon $T$, and the initial condition $\mathbf z_i(0) = \mathbf 0$ for $i \in [n]$, the network error \eqref{eq: NET}  is   bounded as
\begin{align}
\hspace*{-0.08in}
\mathrm{NET} \leq   \sum_{t=1}^T \frac{L^2\alpha_t}{T}   ( 6 \delta^* 
+  9     ), \label{eq: NET_tight_ex}
\end{align}
where $\delta^*  \in \left [1,  \left \lceil \frac{\log{T\sqrt{n}}}{\log{ \sigma_2( \mathbf P_0)^{-1}}} \right \rceil \right ] $  is  the
solution of the optimization problem 
\begin{subequations}
\begin{align}
\displaystyle \minimize_{\beta, \delta} & \quad \delta  \nonumber\\
\st   
& \quad 
\log \beta = \sum_{k={1}}^{\delta-1 } \log  \left ( 1 - \frac{\sum_{i = 1}^{k} u_i b_{i-1} ( \mathbf a_{l_i}^T \mathbf v_{i-1} )^2}{\sigma_2(\mathbf P_0)} \right )  
\label{eq: cons_beta_delta_new}\\
& \quad
   \delta \geq \frac{\log{ T\sqrt{n} }}{\log{ \sigma_2( \mathbf P_0)^{-1}}} -  \frac{\log{\beta^{-1}}}{\log{ \sigma_2( \mathbf P_0)^{-1}}}   \label{eq: beta_ineq_new} \\
&  \quad  
\delta \in \mathbb N_+.
\nonumber
\end{align}\label{eq: prob_delta_new}
\end{subequations} 
\hspace*{-0.1in}
In \eqref{eq: prob_delta_new}, $\beta$ and $\delta$ are optimization variables, for simplicity we replace $\log(x)$ with $\log x$,
and $\mathbb N_+$ denotes the set of positive integers. 

\textbf{Proof:} See Appendix\,\ref{app: net_err}.
\hfill $\blacksquare$
\end{myprop}

Before delving into  Proposition\,\ref{prop: net}, 
we elaborate on the optimization  problem \eqref{eq: prob_delta_new}. 
\begin{itemize}
\item Feasibility: 
Any integer with
 $\delta \geq \frac{\log{T\sqrt{n}}}{\log{ \sigma_2( \mathbf P_0)^{-1}}}  $ is a feasible point to
 problem \eqref{eq: prob_delta_new}.
\item  \textcolor{black}{The variable $\beta$ is {an auxiliary variable that is used to  characterize the temporal variation of $ \sigma_2(\mathbf P_t)$ compared to $\sigma_2(\mathbf P_0)$. It can be eliminated without loss of optimality in solving \eqref{eq: prob_delta_new}}. We refer readers to   \eqref{eq: improvement_ex}\,--\eqref{eq: beta_delta} in Appendix\,\ref{app: net_err}  for more details.}
\item \textcolor{black}{The variable $\delta$ signifies the  mixing time incurred by evolving networks of growing connectivity. And
the constraint  \eqref{eq: beta_ineq_new} characterizes the effect of fast-mixing network with the   error tolerance
$
 \prod_{k={0}}^{\delta-1 } \sigma_2( \mathbf P_k)  \leq \beta \sigma_2( \mathbf P_0)^{\delta} \leq  {1}/{(T\sqrt{n})}
$. }
\item Problem \eqref{eq: prob_delta_new} is a scalar optimization problem with respect to $\delta$. And the optimal $\delta$ is bounded, and can be readily obtained by grid search from $1$ to the integer that satisfies both \eqref{eq: cons_beta_delta_new} and \eqref{eq: beta_ineq_new}.  
\end{itemize}

The solution of problem \eqref{eq: prob_delta_new} is an implicit function of the   edge scheduling scheme $(\{ u_t\}, \{ l_t\})$.  
It is clear form \eqref{eq: cons_beta_delta_new} that
 the larger $u_i b_{i-1} ( \mathbf a_{l_i}^T \mathbf v_{i-1} )^2$ is,  the smaller $\beta$ and  $\delta$ become. This suggests that in order to achieve a better convergence rate, it is desirable to have a faster growth of   network connectivity. 
This is the rational behind using   the edge scheduling method in \eqref{eq: edge_schedule}. 
We provide more insights on Proposition\,\ref{prop: net} as below.

\begin{corollary}
\label{lemma: static_delta}
For a \textit{static} graph with   $u_t = 0$ for $t \in[T]$, the optimal solution of problem \eqref{eq: prob_delta_new} is  given by
 $\delta^* = \left \lceil \frac{\log{T\sqrt{n}}}{\log{ \sigma_2( \mathbf P_0)^{-1}}} \right \rceil$ and $\beta^* = 1$, where $ \left \lceil x \right \rceil $ gives the smallest integer that is greater than $x$.
\end{corollary}
\textbf{Proof:} Since $u_t = 0$ for $t \in[T]$, we have $\beta = 1$.   The optimal value of $\delta$  is then immediately obtained from \eqref{eq: beta_ineq_new}. \hfill $\blacksquare$ 

Corollary\,\ref{lemma: static_delta} states   that problem \eqref{eq: prob_delta_new} has a \textit{closed-form} solution if the underlying graph is  time-invariant. \textcolor{black}{As a result,} Corollary\,\ref{lemma: static_delta} recovers the convergence result in     \cite{ducaga12}.

\begin{corollary}\label{lemma: dist_delta}
Let $(\delta^*, \beta^*)$ be the solution of problem \eqref{eq: prob_delta_new},  for any feasible pair $(\delta, \beta)$ we have
\begin{align}
& \delta^*  -  \left ( \frac{\log{T\sqrt{n}}}{\log{ \sigma_2( \mathbf P_0)^{-1}}} -  \frac{\log{{(\beta^*)}^{-1}}}{\log{ \sigma_2( \mathbf P_0)^{-1}}} \right ) \nonumber \\
\leq  & \delta - \left ( \frac{\log{T\sqrt{n}}}{\log{ \sigma_2( \mathbf P_0)^{-1}}} -  \frac{\log{{\beta}^{-1}}}{\log{ \sigma_2( \mathbf P_0)^{-1}}} \right )
\end{align}
\end{corollary}
\textbf{Proof:}
For $\delta \geq \delta^*$, we have $\beta^* \geq \beta $ from \eqref{eq: cons_beta_delta_new}, and thus obtain
$\delta  \geq \delta^* \geq  \frac{\log{T\sqrt{n}}}{\log{ \sigma_2( \mathbf P_0)^{-1}}} -  \frac{\log{{(\beta^*)}^{-1}}}{\log{ \sigma_2( \mathbf P_0)^{-1}}} 
\geq \frac{\log{T\sqrt{n}}}{\log{ \sigma_2( \mathbf P_0)^{-1}}} -  \frac{\log{{\beta}^{-1}}}{\log{ \sigma_2( \mathbf P_0)^{-1}}}
$. \hfill $\blacksquare$

Corollary\,\ref{lemma: dist_delta} implies that 
 the \textit{minimal} distance between $\delta $ and its lower bound  is achieved at the optimal solution of problem \eqref{eq: prob_delta_new}. Spurred by that, it is reasonable to consider the   approximation
\begin{align}
\delta^* & \approx \left \lceil 
\frac{\log{T\sqrt{n}}}{\log{ \sigma_2( \mathbf P_0)^{-1}}} -  \frac{\log{{(\beta^*)}^{-1}}}{\log{ \sigma_2( \mathbf P_0)^{-1}}}
\right \rceil.
\label{eq: delta_star_appr}
\end{align}
The approximation \eqref{eq: delta_star_appr} facilitates us to  link the   network error \eqref{eq: NET} with the strategy of network topology design. 
Since $\beta^* \leq 1$ (the equality holds when the network is static as described in Corollary\,\ref{lemma: static_delta}),
the use of the evolving network \eqref{eq: Lt_dyn}  reduces the network error. 

\begin{corollary}\label{lemma: schedule}
Let $\{ u_t \}$ and $\{ \hat u_t \}$ be two  edge schedules with
  $u_{t_i} = 1$ and  $\hat u_{\hat t_i} = 1$
for $l_i \in \mathcal F$, where $\mathcal F$ is an edge set,  $l_i$ is the $i$th edge in $\mathcal F$,  
$t_i$ and $\hat t_i$ denote  time steps at which the edge $l_i$ is scheduled under $\{ u_t \}$ and $\{ \hat u_t \}$. 
\[
\text{If $  t_i \leq \hat t_i$, then $  \delta ^* \leq \hat \delta^*$,}
\]
where $  \delta ^*$ and $\hat \delta^*$ are solutions to problem \eqref{eq: prob_delta_new} under $\{ u_t \}$ and $\{ \hat u_t \}$, respectively.
\end{corollary}
\textbf{Proof:} See Appendix\,\ref{app: schedule}.
\hfill $\blacksquare$

Corollary\,\ref{lemma: schedule} suggests that  in order to  achieve a fast mixing  network, it is desirable to add edges   
 as early as possible. 
 
 Combining Proposition\,\ref{prop: net} and Theorem\,1, we present the complete convergence analysis of DDA over evolving networks of growing connectivity in Theorem\,2.
 
{\textbf{Theorem\,2:}} Under the hypotheses of Theorem\,1,   $\psi(\mathbf x^*) \leq R^2$, and $\alpha_t \propto  {R \sqrt{1 - \sigma_2( \mathbf P_0)}}/{( L\sqrt{t})} $,  we obtain for $i \in [n]$
\begin{align}
 f( \hat{\mathbf x}_i(T)) - f(\mathbf x^*) 
= 
\mathcal O \left ( \frac{ RL  \sqrt{1 - \sigma_2( \mathbf P_0)} \delta^*  }{\sqrt{T} } \right )
\label{eq: opt_net_edges}
\end{align}
where 
$\delta^*  \in \left [1,  \left \lceil \frac{\log{T\sqrt{n}}}{\log{ \sigma_2( \mathbf P_0)^{-1}}} \right \rceil \right ] $ is the solution of problem \eqref{eq: prob_delta_new}, and $f = \mathcal O(g)$ means that $f$ is bounded above by  $g $  up to some constant factor.

\textbf{Proof:} See 
Appendix\,\ref{app: rate_DDA}.
\hfill $\blacksquare$

\textcolor{black}{In \eqref{eq: opt_net_edges}, the convergence rate contains two terms: $\mathcal O( \frac{ RL   }{\sqrt{T} } )$ and $\mathcal O( \sqrt{1 - \sigma_2( \mathbf P_0)} \delta^* )$, where the first term holds for most of centralized and decentralized  subgradient methods \cite{nemiyudi83,hazan2016introduction}, and the second term is dependent on the network  topology (in terms of its spectral properties). Here we recall that $1 - \sigma_2( \mathbf P_0) $ is proportional to the network connectivity $\lambda_{n-1}(\mathbf L_0)$, and $\delta^* $  gives
the network mixing time.  
In the special case that  the network is time-invariant, 
 our result reduces to  result in  \cite{{ducaga12}}. In particular, in the time invariant case
  we obtain 
 $\delta^* \approx \frac{\log{T\sqrt{n}}}{1 - \sigma_2( \mathbf P_0)} $ based on Corollary\,1 and Theorem\,2, where
  we have used the fact that  $1 - x \leq \log x^{-1}$ for $x \in (0, 1]$. And thus the dependence on the network topology becomes $ \mathcal O \left ( \frac{  \log(T\sqrt{n})}{ \sqrt{1 - \sigma_2( \mathbf P_0)} } \right )$, leading to the  convergence rate given by  \cite[Theorem\,2]{ducaga12},  $\mathcal O \left ( \frac{RL  \log(T\sqrt{n})}{ \sqrt{T}\sqrt{1 - \sigma_2( \mathbf P_0)} } \right )$. Compared to \cite{ducaga12}, the convergence rate in Theorem\,2 is significantly improved from  $ \mathcal O \left ( \frac{  \log(T\sqrt{n})}{ \sqrt{1 - \sigma_2( \mathbf P_0)} } \right )$ to  $\mathcal O( \sqrt{1 - \sigma_2( \mathbf P_0)} \delta^* )$.  For example, the parameter $\sqrt{1 - \sigma_2( \mathbf P_0)}$ could be quite small when the initial graph $\mathcal G_0$ is sparse.
  In Sec.\,\ref{sec: numerical},  
  we empirically show how the mixing
time $\delta^*$ and the convergence rate improve  when the network connectivity grows. 
 Such  improvement can also be   shown in terms of the convergence time  illustrated in Proposition\,\ref{prop: conv_T_lb}.
}

\begin{myprop}\label{prop: conv_T_lb}
For a dynamic network \eqref{eq: Lt_dyn}, the error bound in \eqref{eq: opt_net_edges} achieves $\epsilon$-accuracy when
the number of iterations of DDA satisfies
\begin{align}
T = \Omega \left ( \frac{1}{\epsilon^2}\frac{1 - \sigma_2(\mathbf P_0)}{\left ( 1 -\alpha \sigma_2( \mathbf P_0) \right)^2}
\right ), \label{eq: T_allpha}
\end{align}
where $f =   \Omega(g)$ means that
$f$ is bounded below by   $g $  up to some constant factor, 
$\alpha = 1 - \frac{\sum_{i = 1}^{\hat K} b_{i-1} ( \mathbf a_{l_i}^T \mathbf v_{i-1} )^2}{\sigma_2(\mathbf P_0)} \geq 0$, and  $\hat K$ is the total  number of added edges based on \eqref{eq: Lt_dyn}.  
\end{myprop}
\textbf{Proof:} 
See Appendix\,\ref{app: time}.
\hfill $\blacksquare$
%

In Proposition\,\ref{prop: conv_T_lb}, the number of iterations   $\Omega(1/\epsilon^2)$ exists due to the nature of the  subgradient method. However, \textcolor{black}{we explicitly show that} the growth of network connectivity, reflected by $\alpha$ in Proposition\,\ref{prop: conv_T_lb}, can affect the convergence of DDA, \textcolor{black}{ which is absent in previous analysis.}
When  $\alpha \to 0$ (corresponding to the case of fast growing connectivity),
 the   number of iterations decreases toward
$T = \Omega( \frac{1-\sigma_2(\mathbf P_0)}{\epsilon^2})$.
This is in contrast with  $T = \Omega( \frac{1}{\epsilon^2}\frac{1}{1 - \sigma_2(\mathbf P_0)} )$ shown in \cite[Proposition\,1]{ducaga12}  when the network is static.

Based on Proposition\,\ref{prop: conv_T_lb}, in Corollary\,\ref{cor: edge_T} we formally state the connection between the edge scheduling  method \eqref{eq: edge_schedule} and  the maximal improvement in the convergence rate of DDA. 
\begin{corollary}\label{cor: edge_T}
The edge scheduling  method \eqref{eq: edge_schedule} provides a greedy solution to minimize the convergence time \eqref{eq: T_allpha}.
\end{corollary}
\textbf{Proof}:
The minimization of  \eqref{eq: T_allpha}  is equivalent to   the minimization of $\alpha$, i.e.,
the       maximization of $\sum_{i} b_{i-1} ( \mathbf a_{l_i}^T \mathbf v_{i-1} )^2$, where the latter is solved by the greedy method in \eqref{eq: edge_schedule}.
\hfill $\blacksquare$


On the other hand, one can understand  Proposition\,\ref{prop: conv_T_lb}
 with the aid of \eqref{eq: single_sigma2t}. Here the latter    yields
\begin{align}
 \sum_{i = 1}^{\hat K}  b_{i-1} (  \mathbf a_{l_i}^T \mathbf v_{i-1}  )^2 \leq  \sigma_{2}(  \mathbf P_{0}  ) - \sigma_{2}(  \mathbf P_{\hat K}  ), \label{eq: alpha_explain}
\end{align}
where $\mathbf P_{\hat K}$ is defined through $\mathbf L_{\hat K}$ as in \eqref{eq: P_t}, and $\mathbf L_{\hat K} = \mathbf L_0 + \sum_{l=1}^{\hat K}  \mathbf a_l \mathbf a_l^T$ given by \eqref{eq: L_new}.
Based on \eqref{eq: alpha_explain}, the       maximization of $\sum_{i = 1}^{\hat K} b_{i-1} ( \mathbf a_{l_i}^T \mathbf v_{i-1} )^2$ in Proposition\,\ref{prop: conv_T_lb} is also linked with  
 the       minimization of $ \sigma_{2}(  \mathbf P_{\hat K}  )$ that is consistent with the edge selection problems \eqref{eq: opt_connectivity}
and \eqref{eq: opt_connectivity_variant}. 
We will empirically verify our established  theoretical convergence results in the next section.

\section{Numerical Results}\label{sec: numerical}

In this section, we demonstrate the effectiveness of the proposed methods \textcolor{black}{via two examples, decentralized $\ell_1$ regression and distributed estimation, where the latter is performed on a real  temperature dataset \cite{chepuri2017learning}.
We first consider   the  $\ell_1$ regression problem. In  \eqref{eq: prob_dist}, let $f_i (\mathbf x) = |y_i - \mathbf b_i^T \mathbf x|$ for $i \in [n]$, where  $\mathcal X = \{ \mathbf x \in \mathbb R^p \, | \, \|\mathbf x \|_2 \leq R \}$ with $p = 5$ and $R = 5$. Here $f_i $   is $L$-Lipschitz continuous with $L = \max_i \| \mathbf b_i \|_2$, and
$\{ y_i \}$ and $\mathbf b_i$  are   data points drawn from the standard normal distribution.}  We   set the initial graph $\mathcal G_0$ as 
a connected random sensor graph  \cite{perparshu14} with $n = 100$ nodes on a unit square; see  an example in Fig.\,\ref{fig: DDA}.  The cost  associated with edge addition is modeled as
$
c_l = \tau_1 e^{\tau_2(d_l - d_0)}  
$, where $d_l$ is the length of the $l$th edge,  $d_0$ is a default communication range during which any two nodes can   communicate  in the low-energy regime, and  $ \tau_1$ and $ \tau_2$ are   scaling parameters. In our numerical examples, we set  $d_0 = 0.7$, $\tau_1 = 10$, and $\tau_2 = 0.5$. In the projected subgradient descent algorithm, we set the step size $\kappa_i = 1/(0.2\sqrt{i})$  and the maximum number of iterations $2000$.
In the decentralized computation of eigenvector (Algorithm\,1), we choose $N_1 = 300$ and $N_2 = 1000$.

\begin{figure}[htb]
\centerline{ \begin{tabular}{cc}
\includegraphics[width=.27\textwidth,height=!]{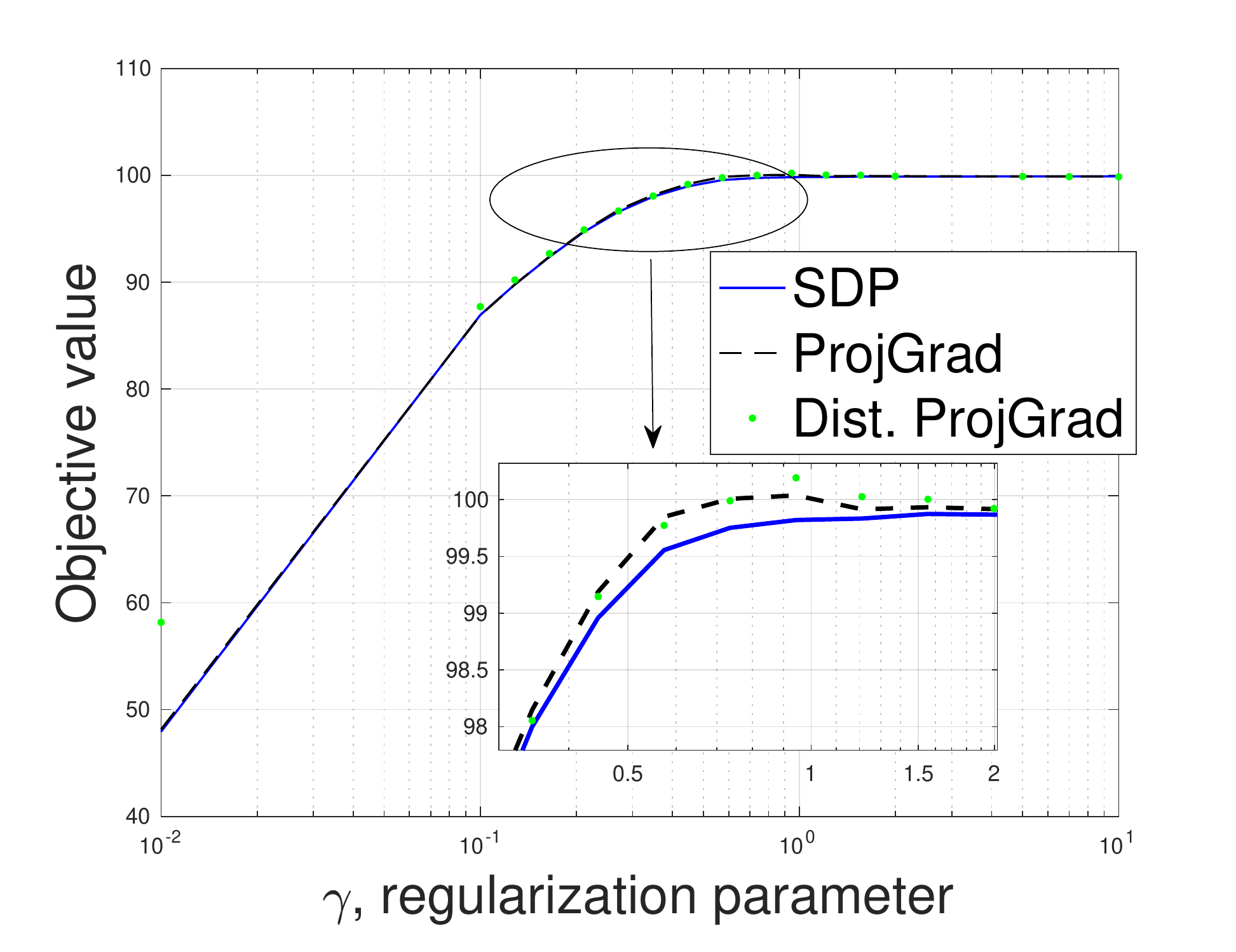}
\hspace*{-0.2in}
&
\includegraphics[width=.27\textwidth,height=!]{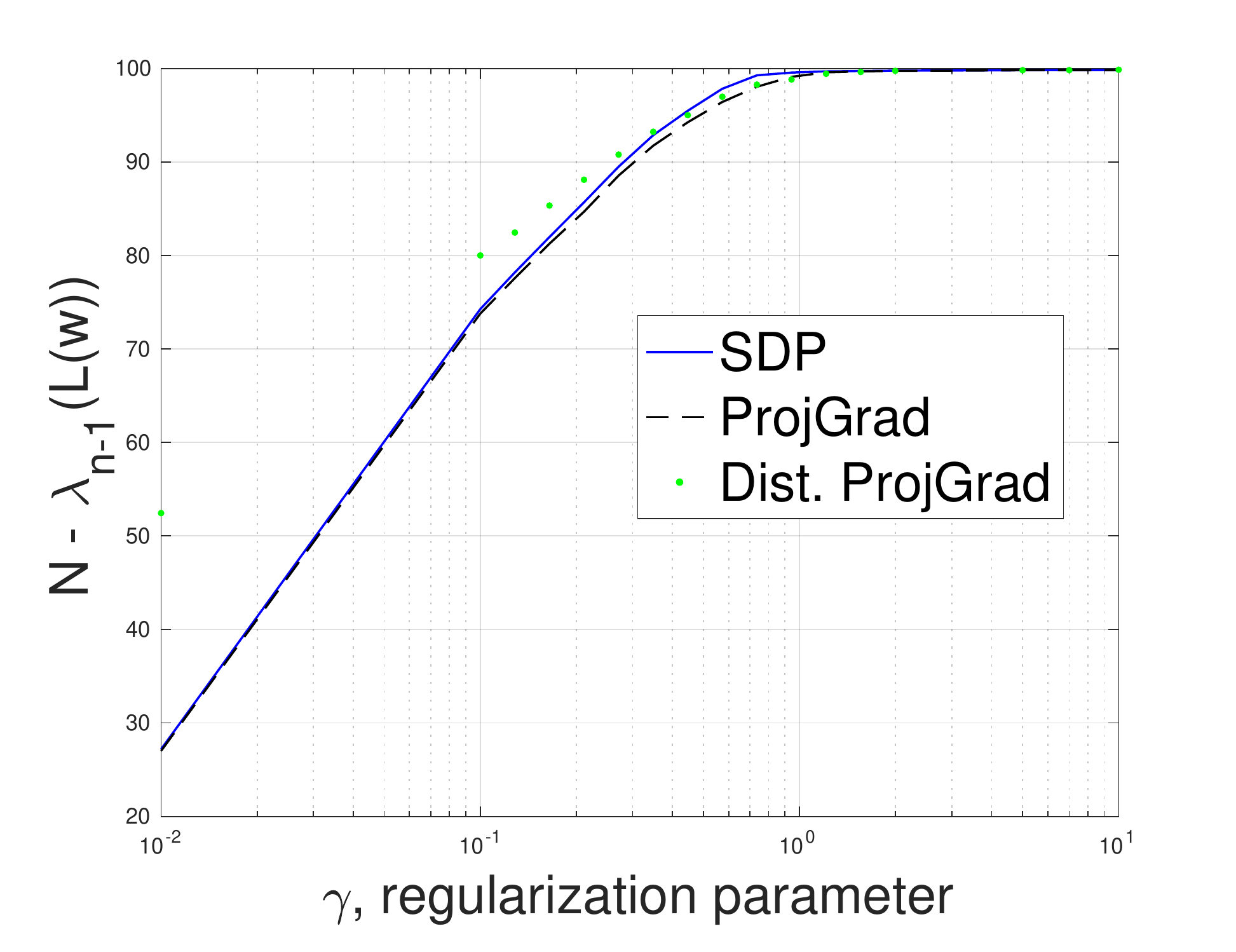}
\\
(a)\hspace*{-0.2in} & (b) \\
\includegraphics[width=.27\textwidth,height=!]{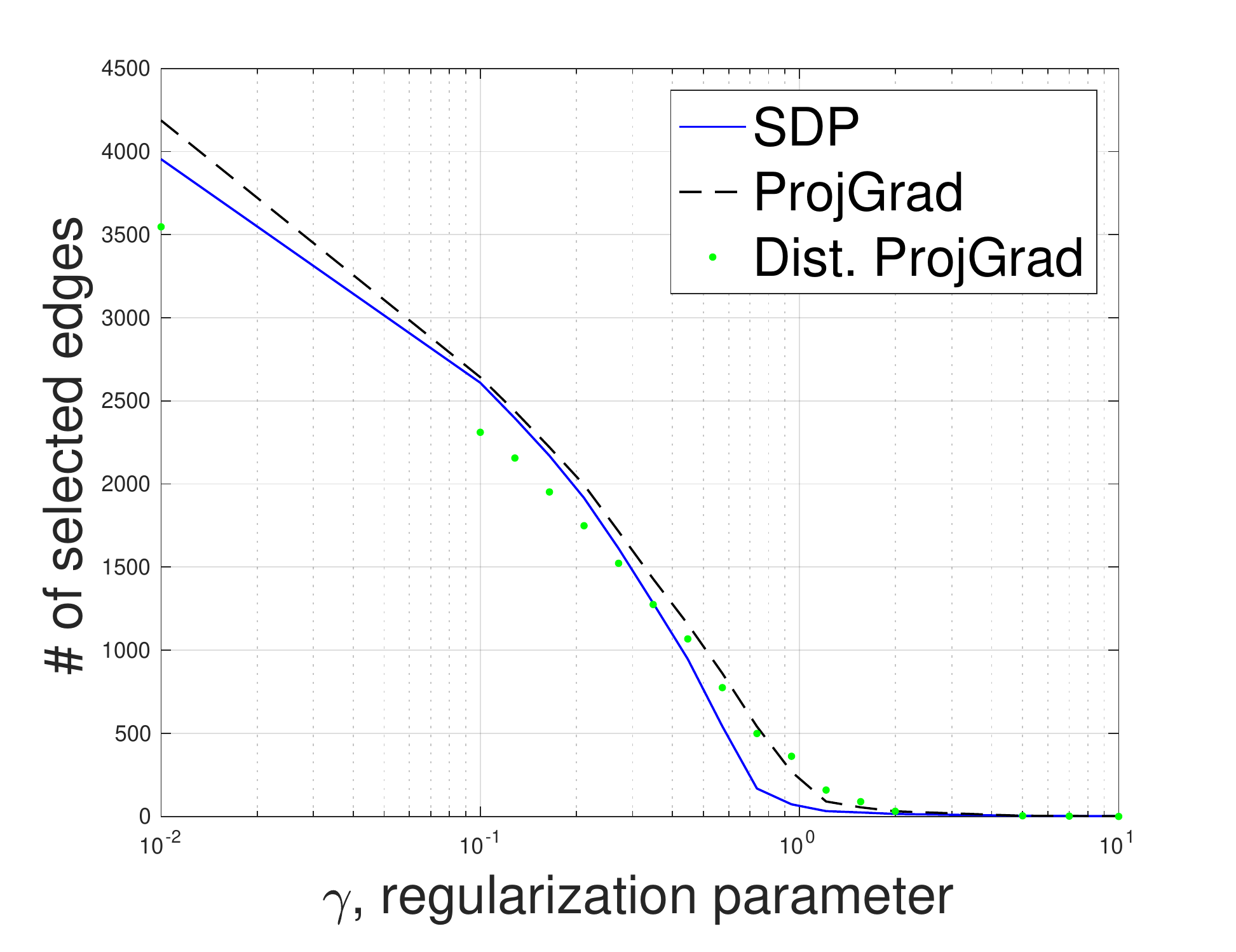}\hspace*{-0.25in}
&
\includegraphics[width=.27\textwidth,height=!] {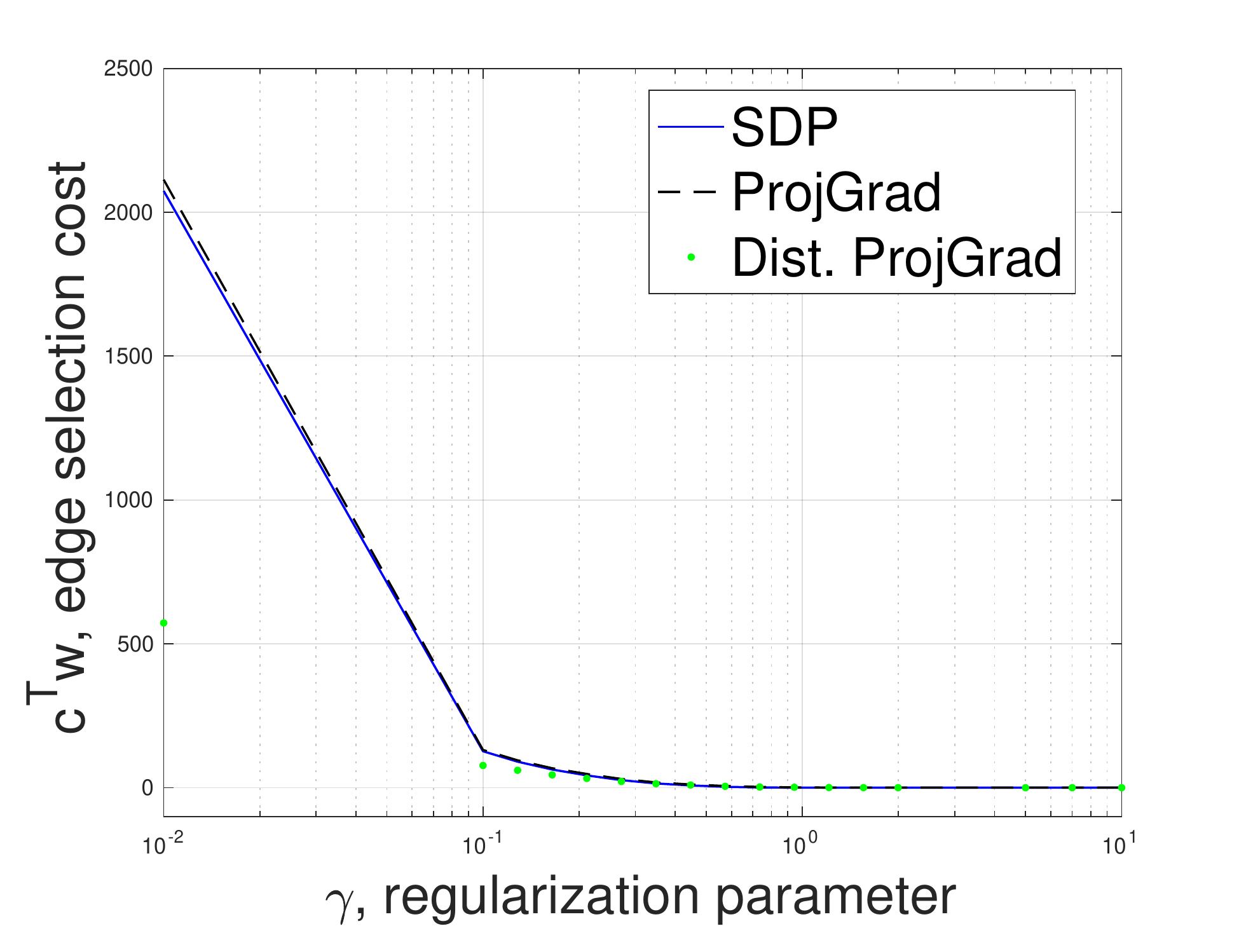}
\\
 (c)\hspace*{-0.25in} & (d)
\end{tabular}}
\caption{\footnotesize{
Solution of problem \eqref{eq: opt_connectivity} obtained from SDP, the projected subgradient algorithm and its distributed variant for different values of regularization parameter $\gamma$:
a) objective value,   b) distance to the maximum algebraic connectivity $n - \lambda_{n-1}(\mathbf L(\mathbf w))$, c) number of selected edges, and  d)
cost of selected edges $\mathbf c^T \mathbf w$.
}}
  \label{fig: obj_gamma}
\end{figure}

\textcolor{black}{
In Fig.\,\ref{fig: obj_gamma}, we present the solution to the problem of network topology design   \eqref{eq: opt_connectivity} as a function of the  regularization parameter $\gamma$, where   problem  \eqref{eq: opt_connectivity} is solved by   SDP,   the projected subgradient algorithm and its distributed variant, respectively.  
In Fig.\,\ref{fig: obj_gamma}-(a), we observe that 
the distributed projected subgradient algorithm  could yield relatively worse optimization accuracy (in terms of  higher objective values) than the other methods. This relative performance degradation occurs because distributed solutions suffer from  
roundoff errors (due to fixed number of    iterations) in average- and max- consensus. 
Compared to SDP,   the advantage of the   subgradient-based algorithms is its low computational complexity, which avoids    the full  eigenvalue decomposition. In Fig.\,\ref{fig: obj_gamma}-(b),  when $\gamma$ increases, the network connectivity decreases, namely, the distance to the maximum algebraic connectivity increases. This is not surprising, since a large $\gamma$ places a large penalty on edge selection so that  the resulting graph is sparse.   In Fig.\,\ref{fig: obj_gamma}-(c),   the number of selected edges, given by the cardinality of the edge selection vector, decreases as $\gamma$ increases.
Further, in Fig.\,\ref{fig: obj_gamma}-(d), the resulting cost of edge selection decreases as $\gamma$ increases.
By varying $\gamma$,
it is clear from Figs.\,\ref{fig: obj_gamma}-(b)-(d) that there exists a tradeoff between the 
network connectivity and the number of edges (or the consumed edge cost). As we can see,  a better network connectivity requires
more edges, increasing the cost.
}

\begin{figure}[htb]
\centerline{ \begin{tabular}{cc}
\includegraphics[width=.27\textwidth,height=!]{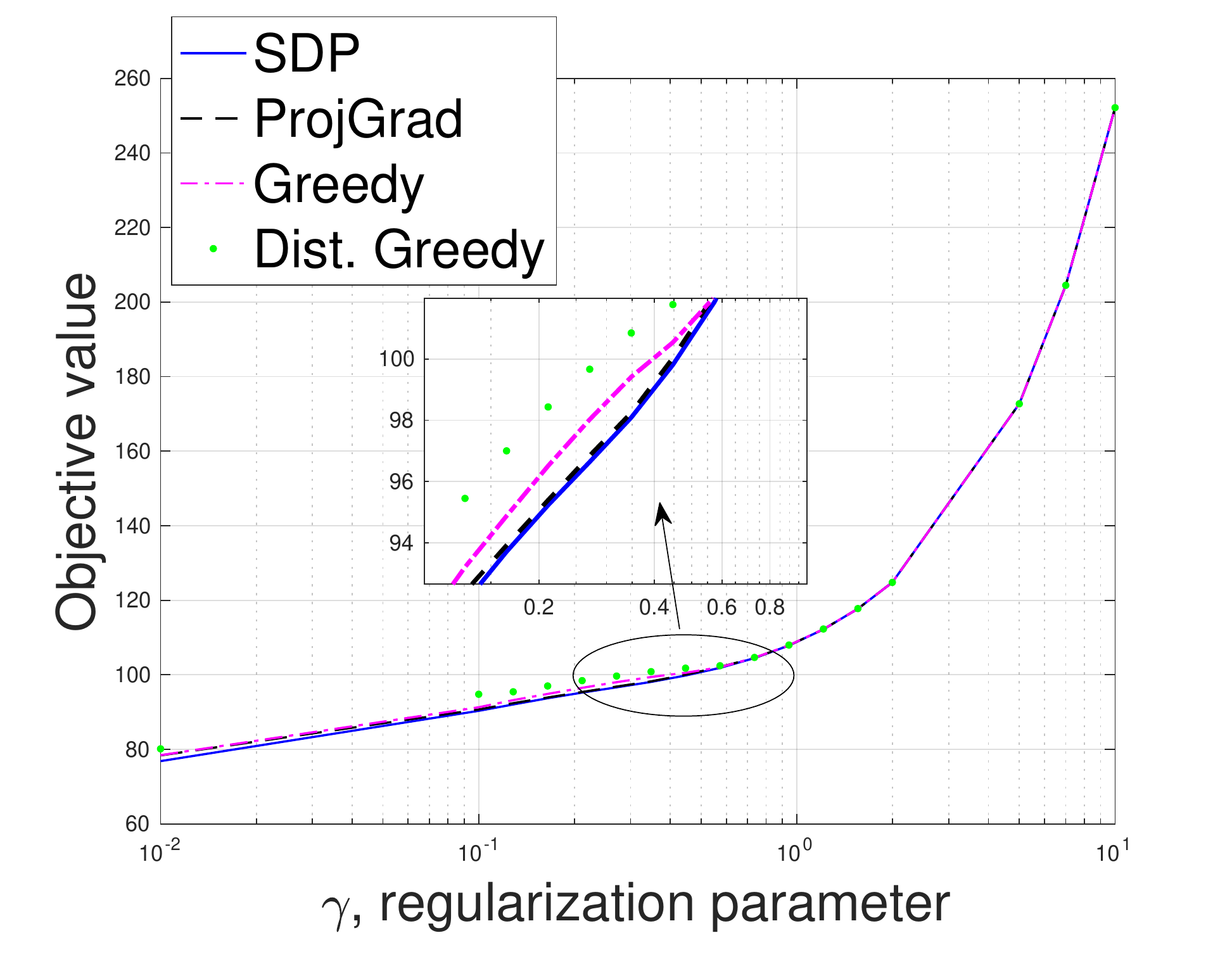}\hspace*{-0.2in}
&
\includegraphics[width=.27\textwidth,height=!]{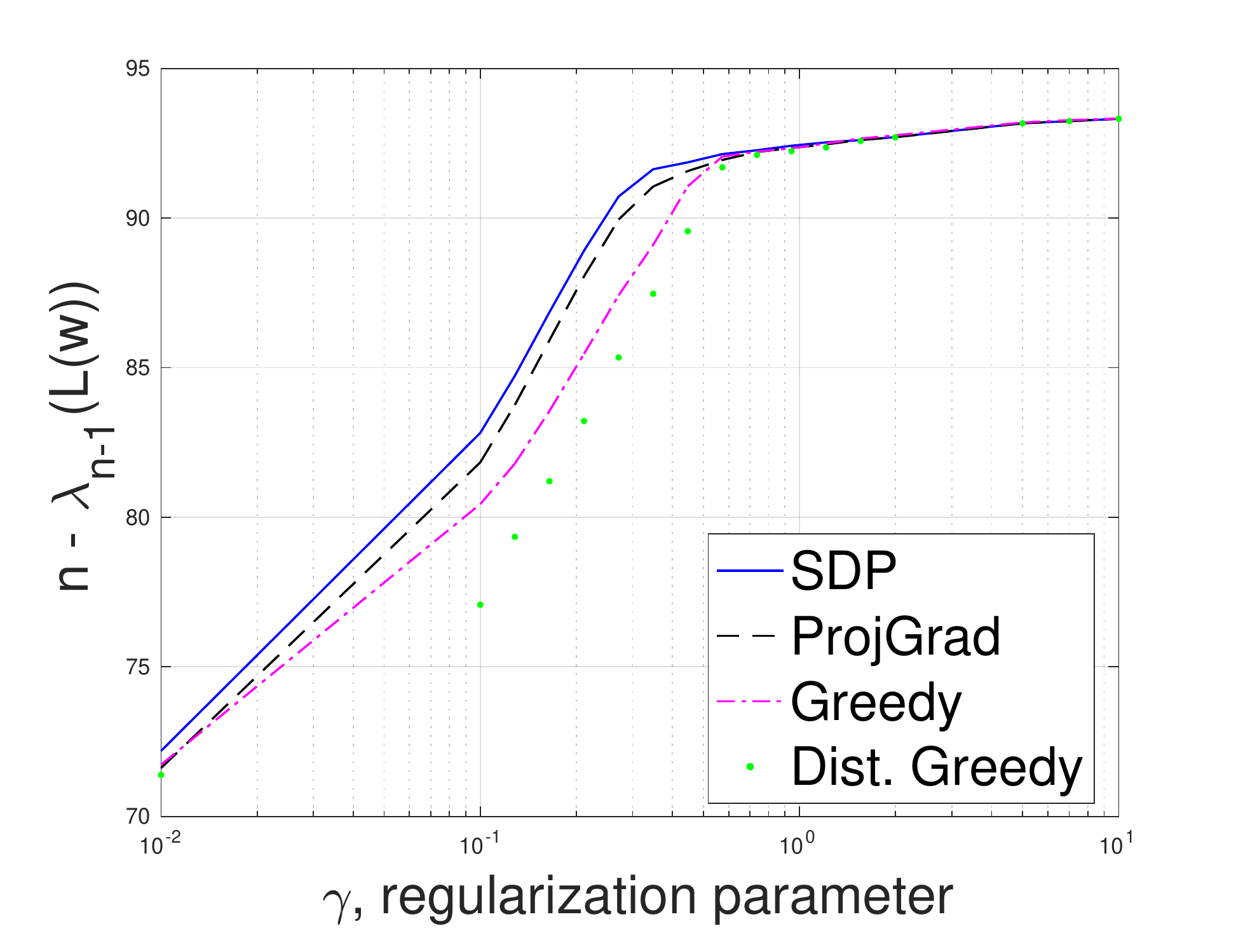}
\\
(a)\hspace*{-0.2in} & (b) \\
\includegraphics[width=.27\textwidth,height=!]{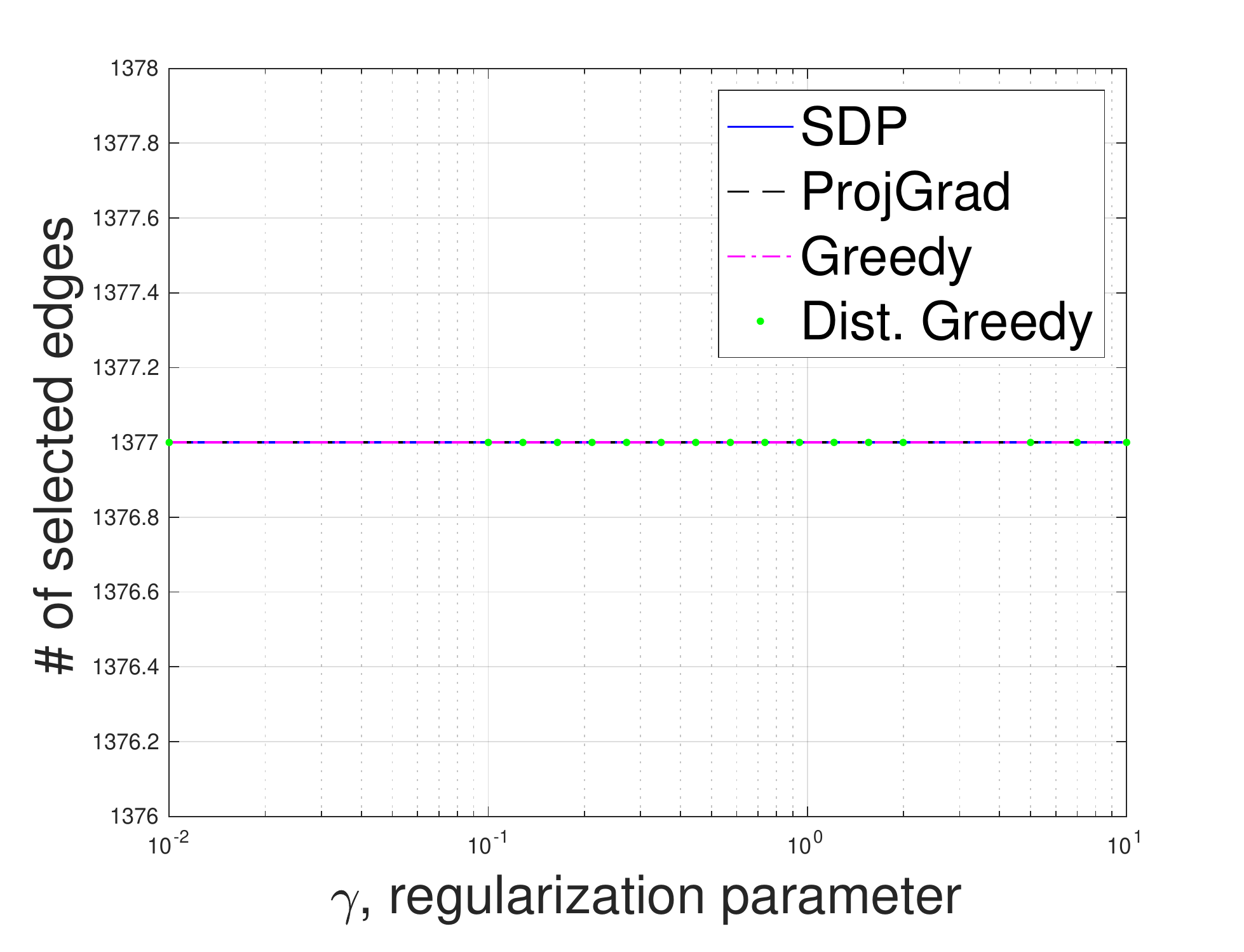}\hspace*{-0.25in}
&
\includegraphics[width=.27\textwidth,height=!]{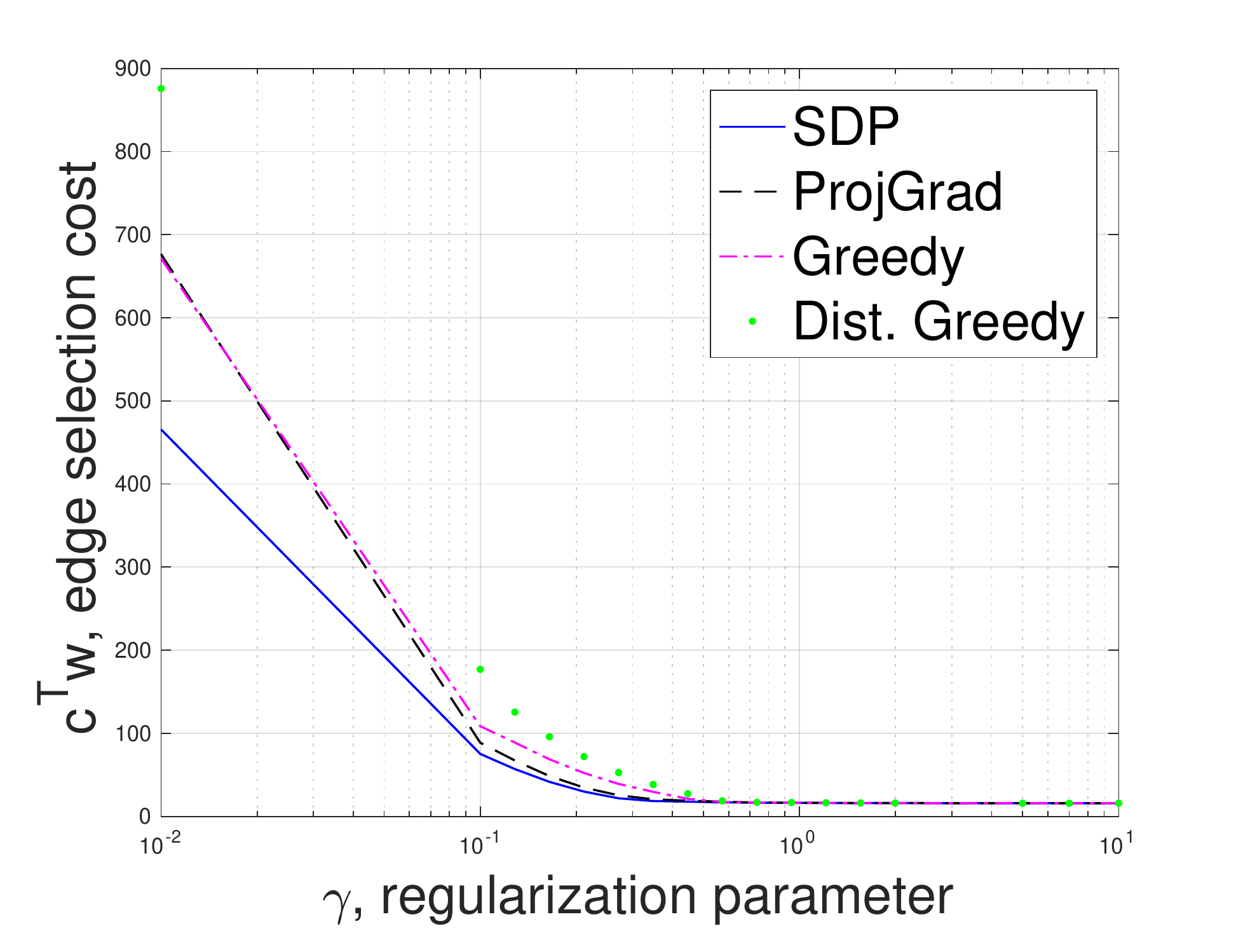}
\\
 (c)\hspace*{-0.25in} & (d)
\end{tabular}}
\caption{\footnotesize{
Solution of problem \eqref{eq: opt_connectivity_variant} obtained from SDP, the projected subgradient algorithm,   the greedy algorithm and its distributed variant for different values of    regularization parameter $\gamma$:
a) objective value,   b) distance to the maximum algebraic connectivity $n - \lambda_{n-1}(\mathbf L(\mathbf w))$, c) number of selected edges, and  d)
cost of selected edges $\mathbf c^T \mathbf w$.
}}
  \label{fig: obj_gamma2}
\end{figure}

\textcolor{black}{
In Fig.\,\ref{fig: obj_gamma2}, we present  the solution to   problem   \eqref{eq: opt_connectivity_variant}, where the number of selected edges is given \textit{a priori}, $k = 1377$.   In Fig.\,\ref{fig: obj_gamma2}-(a), we observe  that the optimization accuracy of the distributed greedy algorithm  is   worse than that of other methods. 
Similar to Fig.\,\ref{fig: obj_gamma}, 
Figs.\,\ref{fig: obj_gamma2}-(b) and (d) show a tradeoff between the  maximum algebraic connectivity and the cost of selected edges.  Fig.\,\ref{fig: obj_gamma2}-(c) 
illustrates the tradeoff mediated by
edge rewiring rather than edge addition  (the number of selected edges is fixed).
}



Next,  we   study the impact of edge selection/scheduling  on the convergence rate of DDA. 
Unless specified otherwise, we employ the projected subgradient algorithm to solve problems   \eqref{eq: opt_connectivity} and \eqref{eq: opt_connectivity_variant}. Given the set of selected edges, 
the dynamic graphical model is given by   \eqref{eq: Lt_dyn}  with 
\begin{align}
u_{i} = \left \{
\begin{array}{ll}
1 & i = (q-1)\Delta +1\\
0 & i = (q-1)\Delta +2 , \ldots, \min\{ q\Delta, T \}
\end{array}
\right. \label{eq: Delta_q}
\end{align}
for $q = 1, 2,\ldots,  \left \lfloor T/\Delta \right \rfloor$, where $ \left \lfloor x  \right \rfloor$ is the largest integer that is smaller than $x$. In \eqref{eq: Delta_q}, $\Delta$ is introduced to represent the topology switching interval. That is, the graph is updated (by adding a new edge) at every $\Delta$ time steps. If $\Delta$ increases, the number of   scheduled  edges would decrease. In an extreme case of $\Delta = T$, the considered graph becomes time-invariant for $t > 1$. 
One can   understand that the smaller   $\Delta$ is, the faster the network connectivity grows.
In \eqref{eq: Delta_q}, we also note that one     edge   is    added at the beginning of each time interval, which is motivated by Corollary\,\ref{lemma: schedule} to add edges as early as possible. 

\begin{figure}[htb]
\centerline{ \begin{tabular}{cc}
\includegraphics[width=.225\textwidth,height=!]{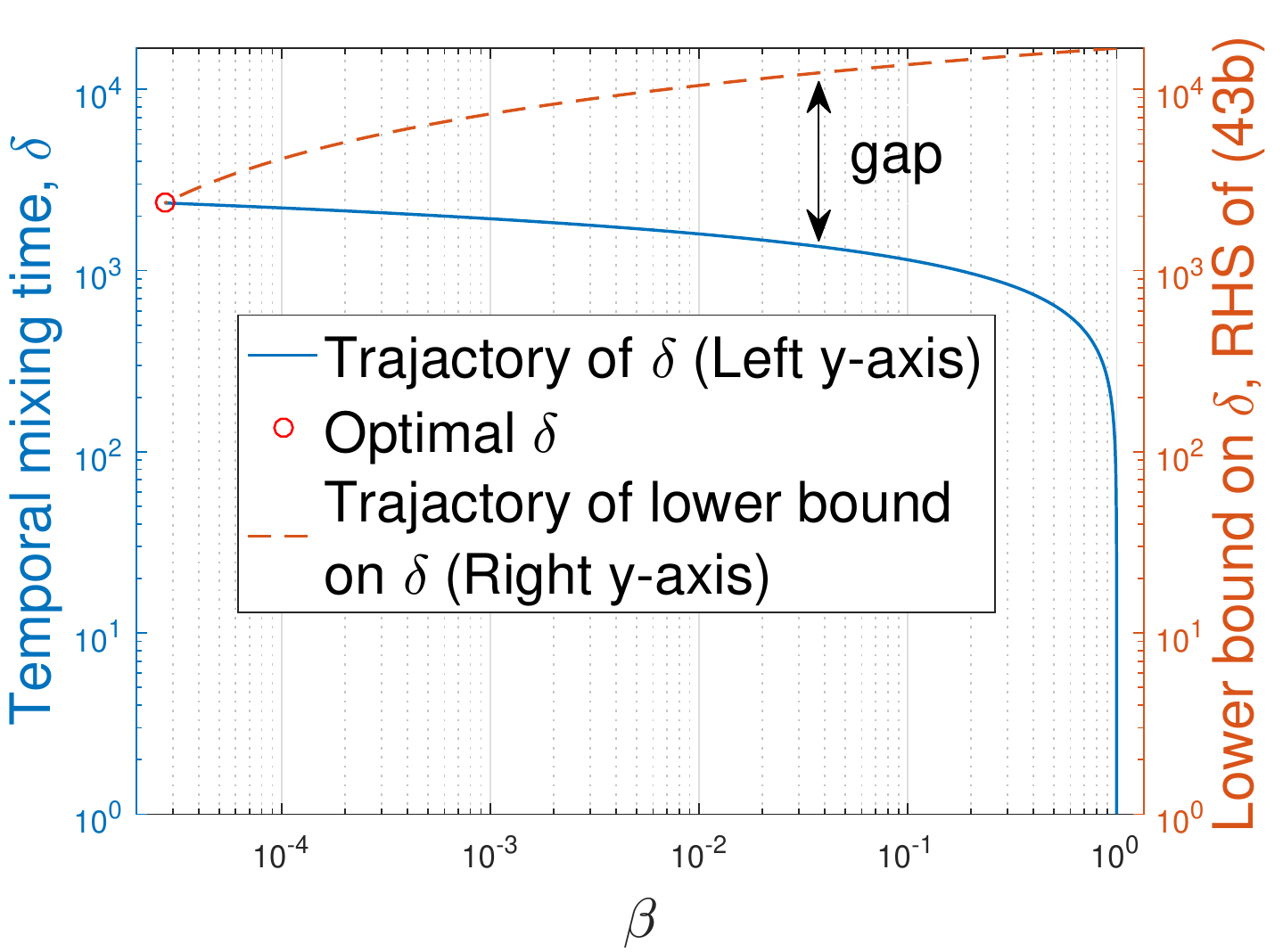}
& \hspace*{-0.15in} \includegraphics[width=.245\textwidth,height=!]{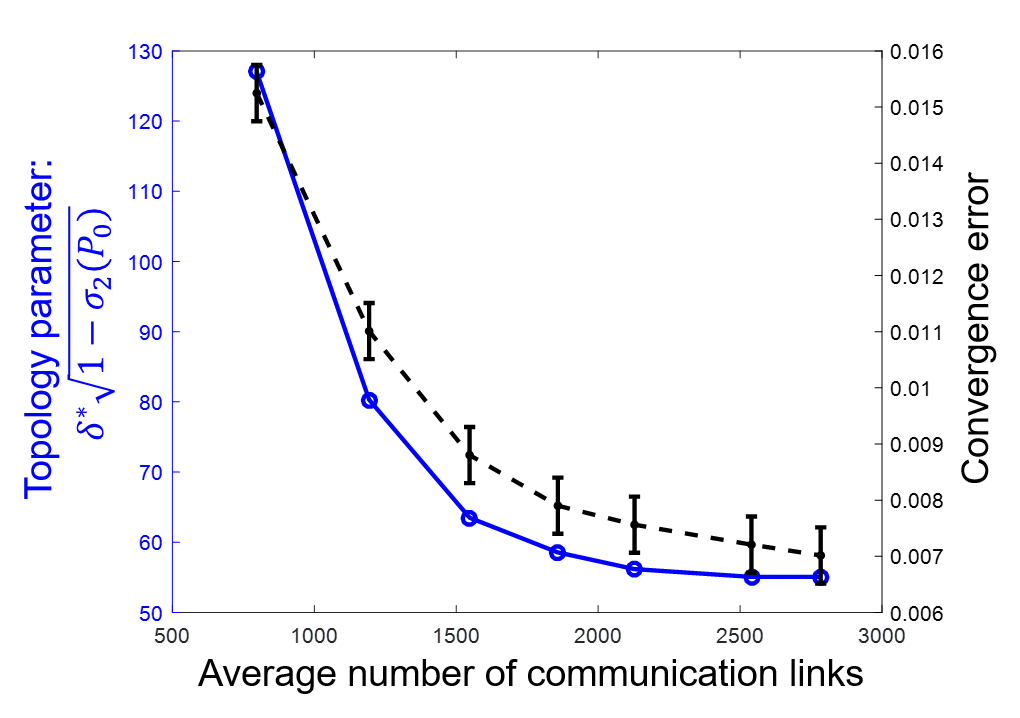}
\\
(a) & \hspace*{-0.15in} (b) 
\end{tabular}}
\caption{\footnotesize{ Illustration of the optimal mixing time: a)
solution path against $\beta$, \textcolor{black}{b)   mixing time and convergence error (regret) versus number of edges (communication links).}
}}
  \label{fig: delta}
\end{figure}

In Fig.\,\ref{fig: delta}, we demonstrate  the optimal temporal mixing time $\delta$ in \eqref{eq: prob_delta_new},
where $\Delta =1$, $T = 50000$, and we set $\gamma = 0.01$ and $k = 1000$ while solving \eqref{eq: opt_connectivity_variant}. In Fig.\,\ref{fig: delta}-(a), we present the   solution path to problem \eqref{eq: prob_delta_new}.  We recall from \eqref{eq: prob_delta_new} that the optimal $\delta$ is obtained by searching positive integers in  an ascending order until the  pair $(\beta, \delta)$ satisfies \eqref{eq: beta_ineq_new}, where 
$\beta$  is a function of $\delta$ given by \eqref{eq: cons_beta_delta_new}. As we can see, at the beginning   of the integer search, a small $\delta$ corresponds to a large  $\beta$, which leads to a large  lower bound on $\delta$, and thus violates the constraint \eqref{eq: beta_ineq_new}. As the value of $\delta$ increases, the value of  the lower bound   decreases since $\beta$ decreases.  This procedure terminates until   
\eqref{eq: beta_ineq_new} is satisfied, namely, the circle point  in the figures. We   observe that  the lower bound on $\delta$ is quite close to the optimal $\delta$, which validates  the approximation in \eqref{eq: delta_star_appr}. 

\textcolor{black}{In Fig.\,\ref{fig: delta}-(b), we   show how the mixing time and the   convergence error (also known as regret) improves when the number of edges  (communication links) increases by varying $k$ in \eqref{eq: opt_connectivity_variant}. 
Here the regret is measured empirically as  $\max_i [f(\hat {\mathbf x}_i(T)) - f(\mathbf x^*)]$, where $\mathbf x^*$ denotes the optimal centralized solution. In the plots, the empirical function error is averaged over $20$ numerical  trials, and the error bar  denotes one standard error.
We observe that  both the mixing time and the regret can be significantly improved as the number of communication links increases. This indicates  the importance of the dynamic  network topology in accelerating the convergence   of DDA.
Also, the improvement trend of the regret is similar  to the mixing time. This is consistent with the theoretical predictions of Theorem\,2.
Furthermore, the regret   ceases  to improve  when the graph has attained $50\%$ sparsity, i.e., it has half the edges of the complete graph. This suggests that significant savings in computation and communication can be attained without significant effect on performance.
}

\begin{figure}[htb]
\centerline{ \begin{tabular}{c}
\includegraphics[width=.45\textwidth,height=!]{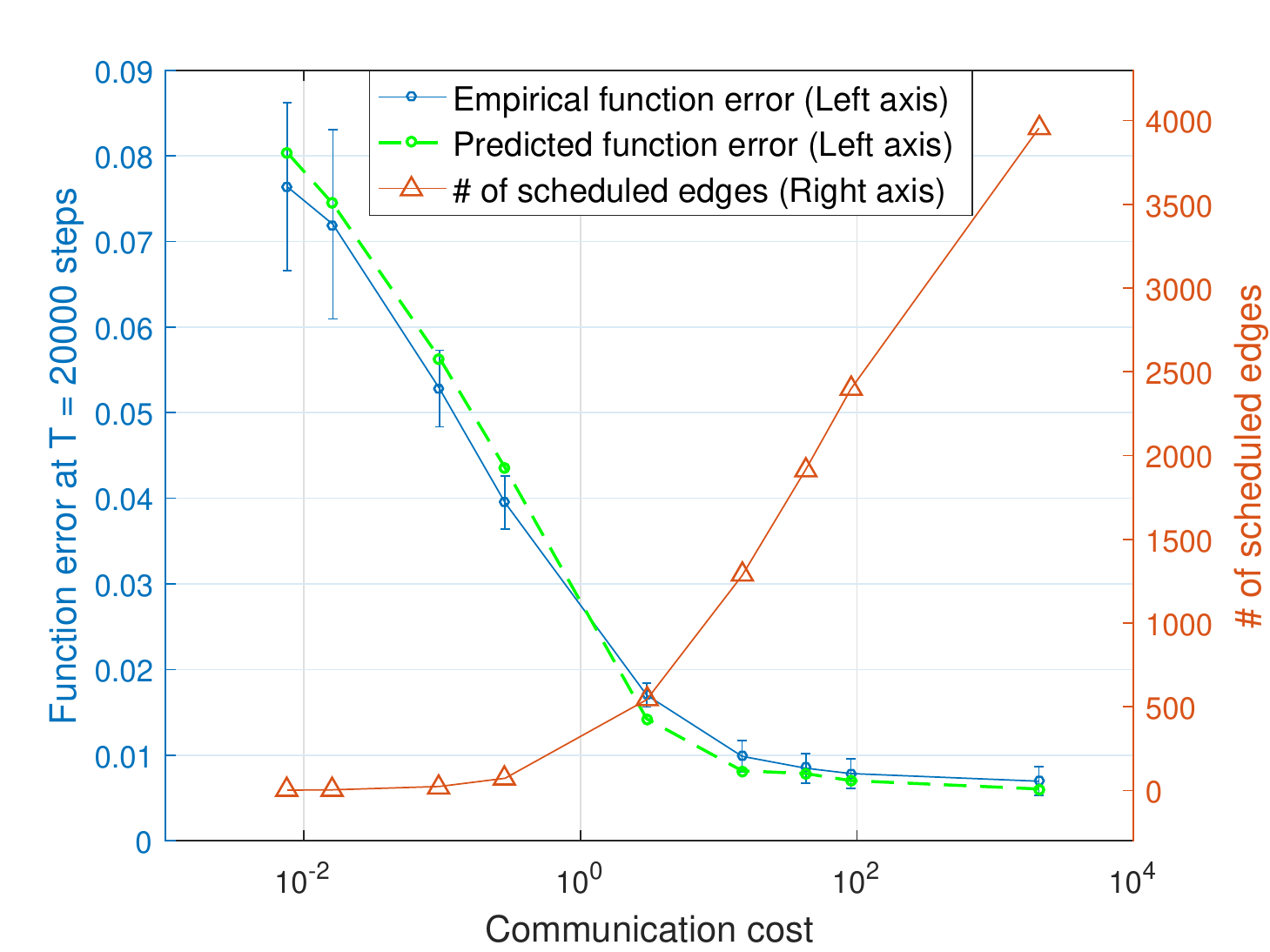}
\\
(a) \\
\hspace*{-0.15in} \includegraphics[width=.45\textwidth,height=!]{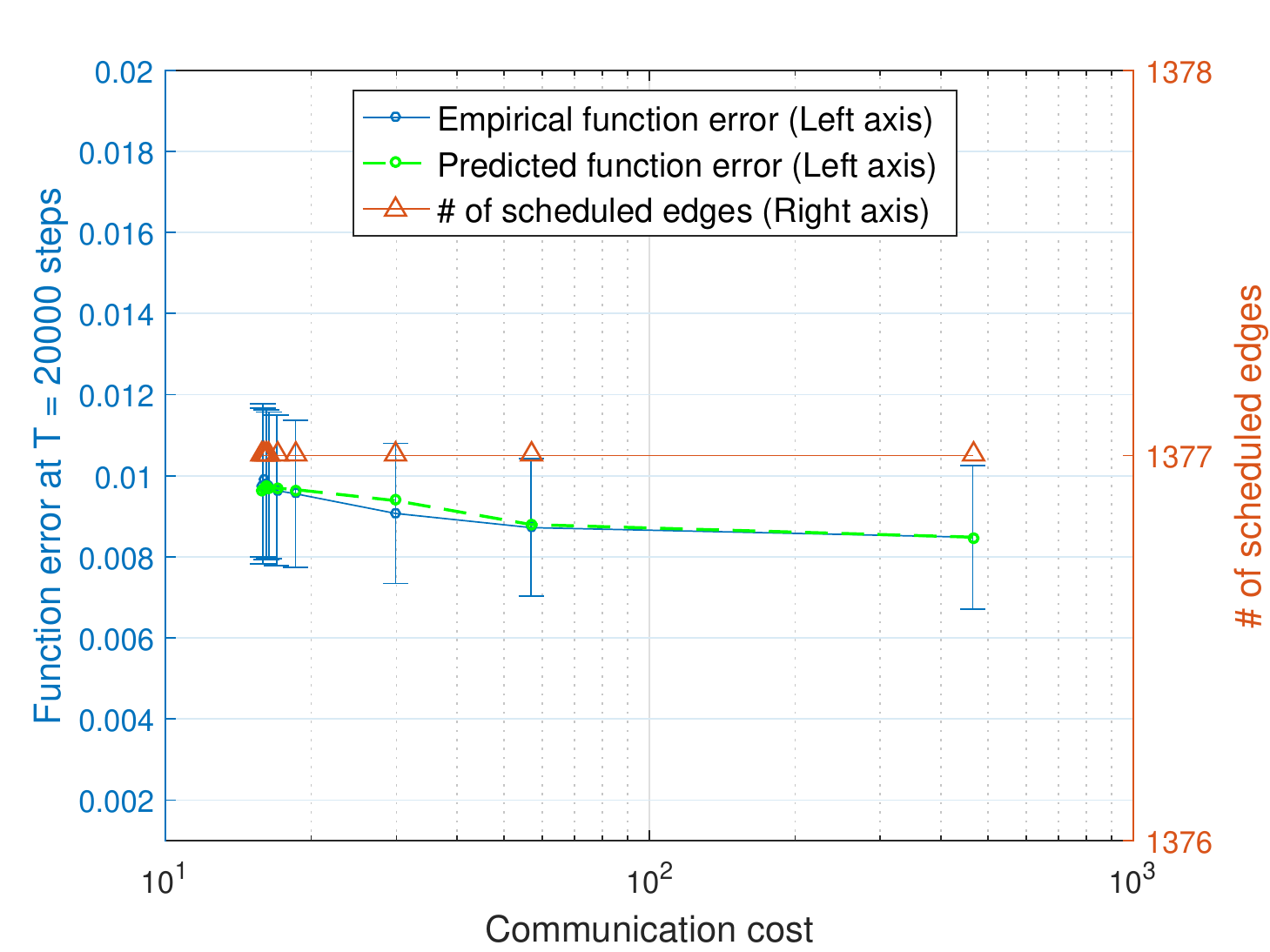}
\\
(b) 
\end{tabular}}
\caption{\footnotesize{
Function error at $T = 20000$ versus the communication cost when $\Delta = 1$: a) edge selection scheme provided by the solution of 
problem    \eqref{eq: opt_connectivity}; b) edge selection scheme provided by the solution of 
problem    \eqref{eq: opt_connectivity_variant}.
}}
  \label{fig: Err_gamma}
\end{figure}

In Fig.\,\ref{fig: Err_gamma}, we present the regret  versus the communication cost consumed during DDA with $T = 20000$ and  $\Delta =1 $.  
For  comparison, we   present   the   predicted function error   in Theorem\,2 that is scaled up to constant factor. 
As we can see, there is a good agreement between the empirical function error  and the theoretical prediction. 
Moreover, we   show the number of added edges versus the communication cost consumed during DDA. 
\textcolor{black}{
In Fig.\,\ref{fig: Err_gamma}-(a), we 
show the behavior of the solution  of
  problem \eqref{eq: opt_connectivity}  for different values of $\gamma$.
Note that, consistent with consistent with Fig.\,\ref{fig: delta}-(b),
the value of adding more edges diminishes when the total number of edges increases above the $50\%$ sparsity level.  
}

 In Fig.\,\ref{fig: Err_gamma}-(b), we present
 the function error by  solving problem   \eqref{eq: opt_connectivity_variant} under a fixed number of selected edges, $k = 1377$.
Compared to  Fig.\,\ref{fig: Err_gamma}-(a),  the communication cost increases  due to edge rewiring rather than edge addition.
However, 
 the improvement on the function error of DDA is less significant. This implies that the number of scheduled edges plays a key role on the acceleration of DDA.

\begin{figure}[htb]
\centerline{ \begin{tabular}{c}
\includegraphics[width=.45\textwidth,height=!]{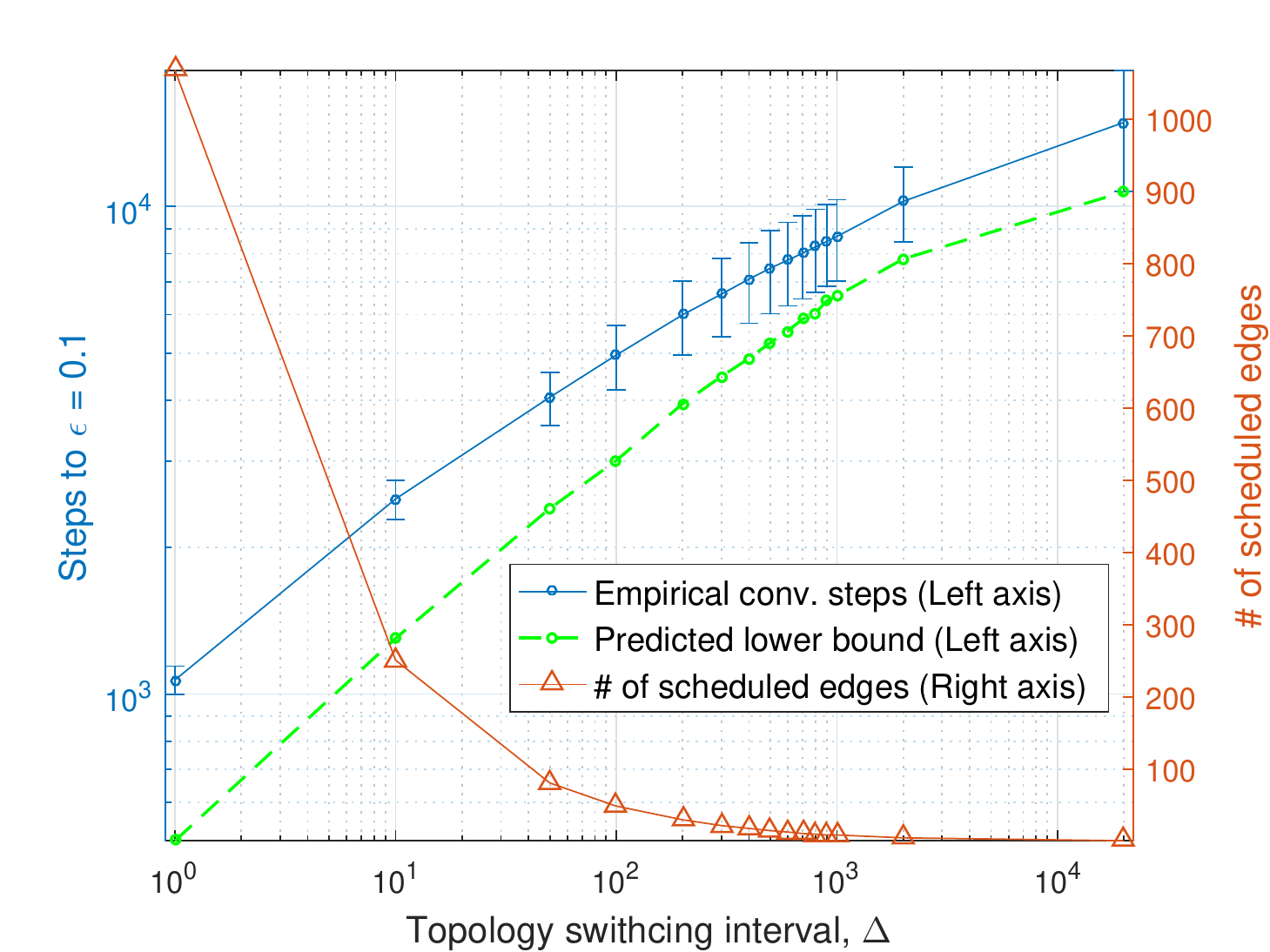}
\end{tabular}}
\caption{\footnotesize{
Convergence steps to $\epsilon$-accuracy solution versus topology switching interval $\Delta$.   
}}
  \label{fig: T_Delta}
\end{figure}

In Fig.\,\ref{fig: T_Delta}, we present the empirical convergence time of DDA for different values of topology switching interval $\Delta$ in \eqref{eq: Delta_q}. Here the empirical convergence time is averaged over $20$ trials, the  accuracy  tolerance is chosen as $\epsilon = 0.1$, and  the edge selection scheme is given by the solution
problem   \eqref{eq: opt_connectivity} at $\gamma = 0.01$.
For comparison, we  plot the theoretical prediction of the lower bound on the convergence time in   Proposition\,\ref{prop: conv_T_lb}, and  the number of scheduled edges  while performing DDA. 
As we can see, the convergence time increases as $\Delta$ increases since the network connectivity grows faster, evidenced by the increase of scheduled edges. Moreover, we observe that the convergence behavior of DDA is improved significantly even under a relatively large $\Delta$, e.g., $\Delta = 100$, compared to that of using a static network at $\Delta = 20000$. 
This   result 
 implies that one can  accelerate the convergence   of DDA even in the regime of low switching rate through network topology design.
Lastly, we note that  the  variation of the empirical  convergence behavior is predicted well by our theoretical results in Proposition\,\ref{prop: conv_T_lb}.

%

\subsection*{Real-world application: distributed estimation over a sensor network}
\textcolor{black}{
In what follows,  we consider an application of distributed estimation based on temperature data collected across $32$ weather  stations at $K = 10$ sampling points \cite{chepuri2017learning}. 
The estimation problem is given by
\cite[Sec.\,IV-C]{HosChaMes16} 
\begin{align}\label{eq: dist_est}
\begin{array}{ll}
\displaystyle \minimize_{\mathbf x} & \frac{1}{n} \sum_{i=1}^n \left ( \sum_{t=1}^K \| y_{i,t} - \mathbf b_i^T \mathbf x\|_2^2 \right ) + \mathcal I (\mathbf x),
\end{array}
\end{align}
where $\mathbf x$ denotes  the vector of temperature intensities to be estimated at some points in space,  
$y_{i,t} $ is the temperature observation at station   $i$ and  time $t$, 
$\{ \mathbf b_i \}$ are 
  observation vectors inferred from the past sensor observations and parameter estimate using 
    the   method in \cite{chepuri2017learning}, and $\mathcal I (\mathbf x)$ is an indicator function,
$\mathcal I (\mathbf x) = 0$ if $\| \mathbf x \|_2 \leq 5$, and $0$ otherwise. Since problem \eqref{eq: dist_est} contains the particular   smooth+nonsmooth composite objective function,  
 a proximal gradient exact first-order algorithm (PG-EXTRA) \cite{shi2015proximal}  and a distributed linearized alternating direction method of multipliers (DL-ADMM) \cite{aybat2017distributed}  can be used for distributed estimation,
 with a linear convergence rate 
   faster than DDA.  Note that PG-EXTRA and DL-ADMM were defined under static networks while our approach is proposed for evolving networks of growing connectivity. In a fair comparison, 
   we  determine an equivalent-static  network by solving the network design problem \eqref{eq: opt_connectivity_variant}, so that the number of edges of this static network is the same as the average number of edges of the proposed time-evolving network. 
} 

\begin{figure}[htb]
\centerline{ \begin{tabular}{c}
\includegraphics[width=.45\textwidth,height=!]{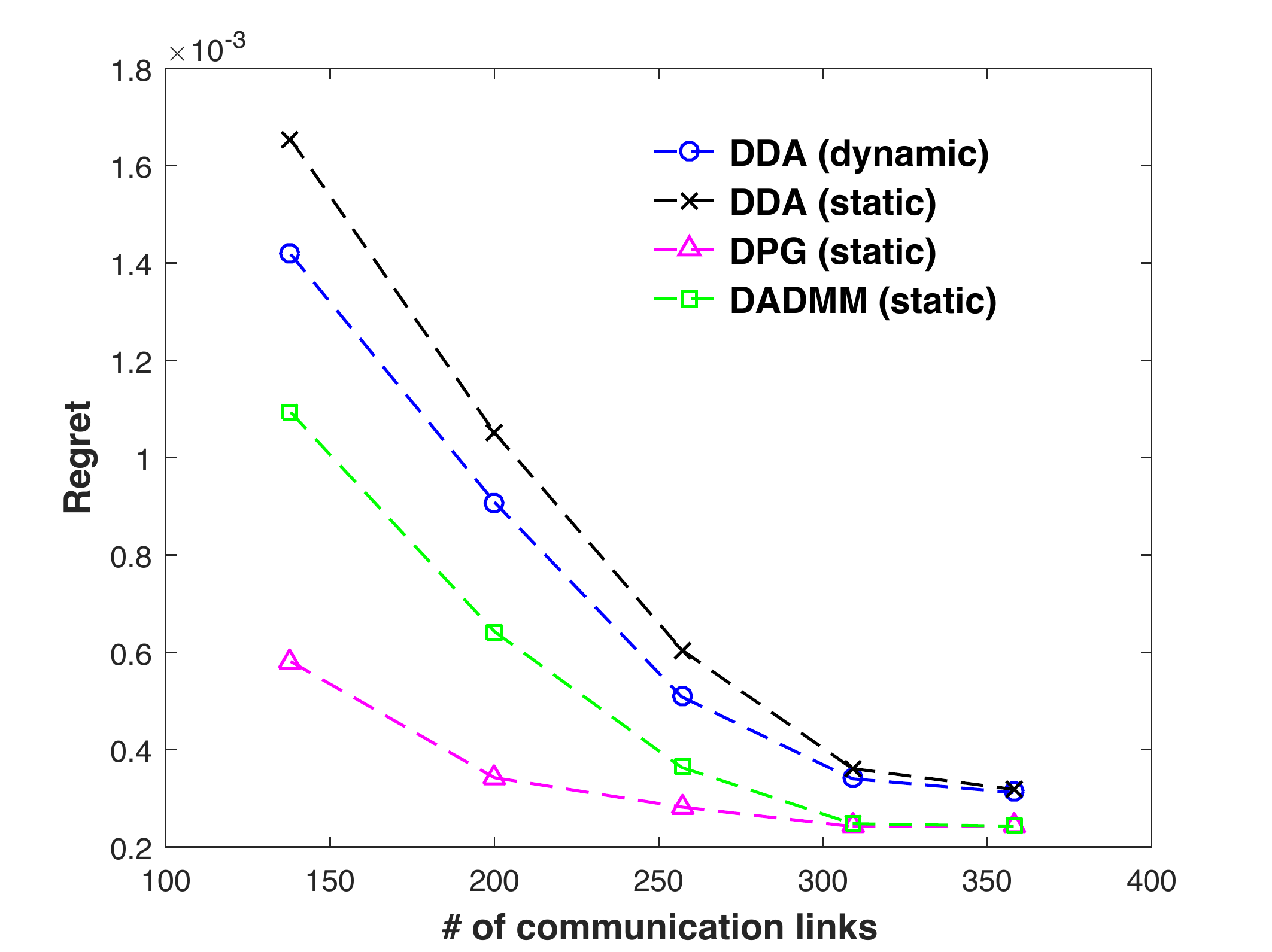}
\end{tabular}}
\caption{\footnotesize{
Regret against average number of communication links.
}}
  \label{fig: mse}
\end{figure}

\begin{figure}[htb]
\centerline{ \begin{tabular}{cc}
\includegraphics[width=.25\textwidth,height=!]{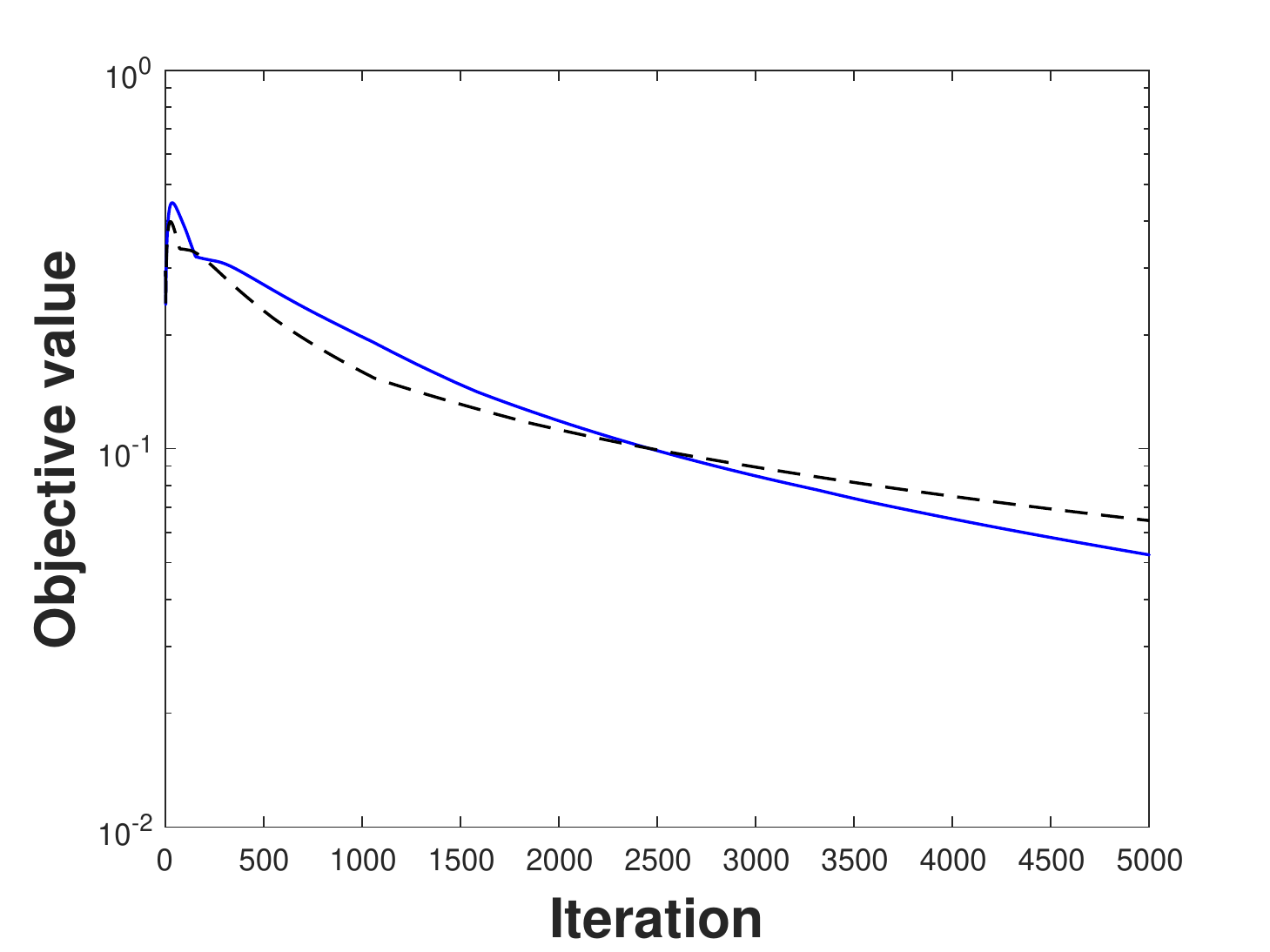}
&  \includegraphics[width=.25\textwidth,height=!]{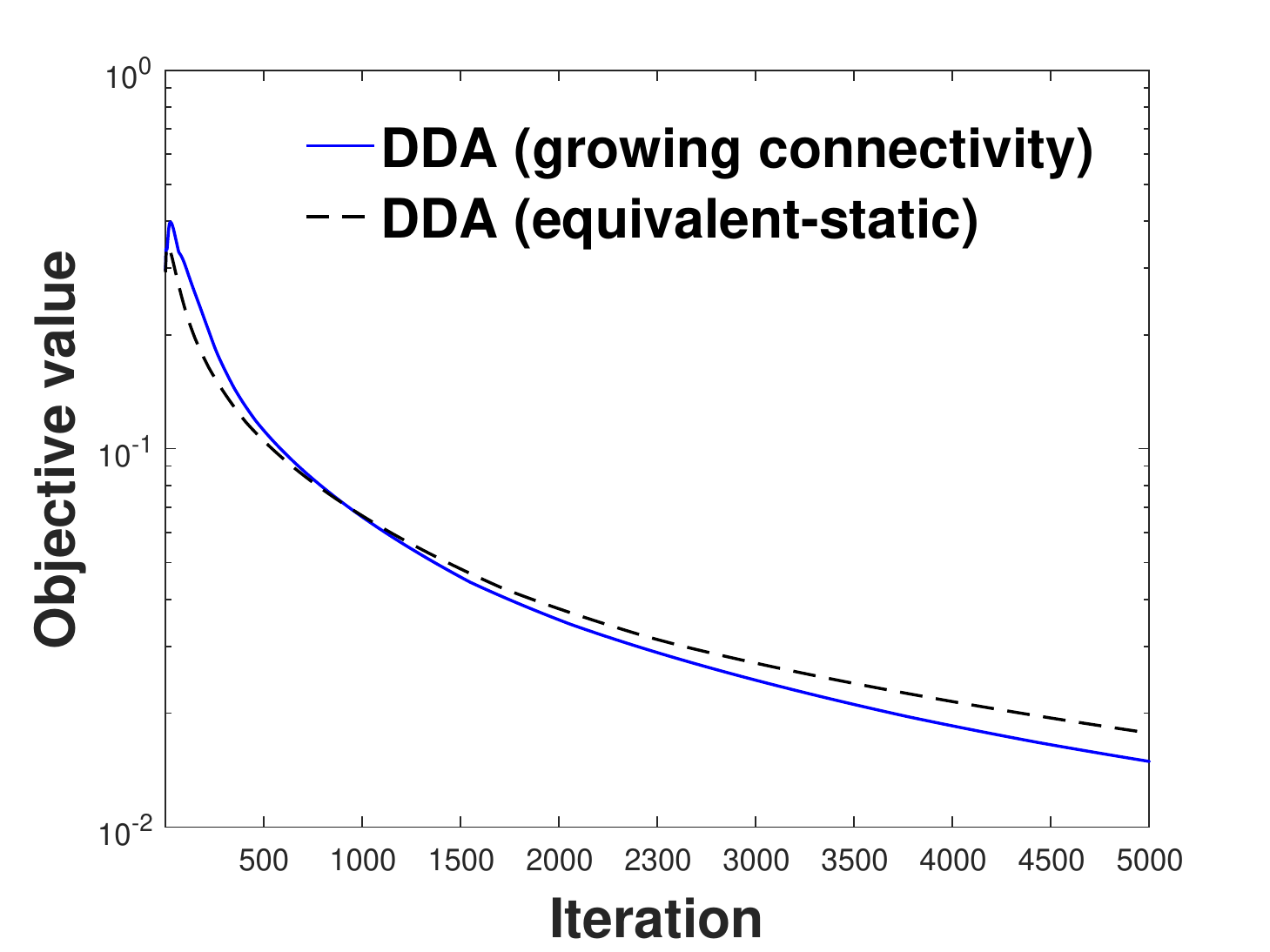}
\\
(a) & \hspace*{-0.15in} (b) 
\end{tabular}}
\caption{\footnotesize{Convergence trajectory at different number of communication links: a) 138 and b) 200.
}}
  \label{fig: dist_iter}
\end{figure}

\textcolor{black}{
   In Fig.\,\ref{fig: mse}, we present the limiting regret
   as a function of the average number of  communication links utilized by DDA. As we can see,  PG-EXTRA and DL-ADMM outperform the DDA-based algorithms.
    This is not surprising, since both
PG-EXTRA and L-ADMM  have a linear convergence rate, faster than DDA. Specifically, the former  utilizes historical information to obtain a 
better 
gradient estimate, and the latter uses the operator splitting method to accelerate the convergence rate. However, the performance gap between DDA and PG-EXTRA (or DL-ADMM) decreases as the number of communication links increases. 
We further note that when applied to the time varying network topology,  DDA has better   performance than that of using  the equivalent-static   network. 
In Fig.\,\ref{fig: dist_iter}, we compare   convergence trajectories when the avarage number of edges is    $138$ and $200$, respectively. As we can see, 
the convergence speed is accelerated  when the number of communication links increases.
\textcolor{black}{We also note that our approach converges slower at the beginning, however, it accelerates after a certain number of iterations due to the  growing network connectivity. This leads to a $18\%$   improvement in   the objective function at the end of the time horizon.}
}

\section{Conclusions}\label{sec: conclusion}
In this paper, we   studied the DDA algorithm for distributed convex optimization in multi-agent networks.
We took into account the impact of network topology design on the convergence analysis of DDA. 
We showed that the acceleration of DDA is achievable under   evolving networks of growing connectivity, where the growing networks can be designed via  edge selection and scheduling.
We  
demonstrated
the tight connection between the  improvement in convergence rate and the growth speed of network connectivity,  which is absent in the existing analysis. 
Numerical results showed that 
our  theoretical predictions are well matched to the empirical convergence behavior of accelerated DDA.
There are multiple directions for future research. We would
like to 
study the convergence rate of DDA when the network connectivity grows in average but not
monotonically.
It will also
be of interest to address the issues of communication delays and link failures in the accelerated DDA algorithm. 
Last but not the least,  we would  relax the assumption of     deterministically time-varying networks to randomly varying networks. 
\bibliographystyle{IEEEbib}
\bibliography{journal_col}

\appendices

\section{Proof of Proposition\,\ref{prop: inc_connect}}\label{app: inc_connect}
If $u_t = 0$, we obtain $\lambda_{n-1}(\mathbf L_t) = \lambda_{n-1}( \mathbf L_{t-1} ) $ from \eqref{eq: Lt_dyn}. In what follows, we consider the case of $u_t = 1$.
Let $d$ denote a lower bound on $\lambda_{n-1}(\mathbf L_t) $, namely, $ d \leq \lambda_{n-1}(\mathbf L_t)$. By setting
$
\epsilon \Def d - \lambda_{n-1}( \mathbf L_{t-1} )
$, it has been shown in \cite[Eq.\,12]{ghoboy06} that the desired $\epsilon$ satisfies
\begin{align}
\frac{c^2}{\epsilon} \geq 1+ \frac{2}{\delta - \epsilon}, \label{eq: key_eq}
\end{align}
where $c \Def  \mathbf a_{l_t}^T \mathbf v_{t-1} $, 
$\delta \Def \lambda_{n-2}(\mathbf L_{t-1}) - \lambda_{n-1}(\mathbf L_{t-1})$, and $\delta - \epsilon =\lambda_{n-2}(\mathbf L_{t-1})  - d \geq  \lambda_{n-2}(\mathbf L_{t-1})  - \lambda_{n-1}(\mathbf L_t) >0$.

Multiplying \eqref{eq: key_eq} by $\epsilon(\delta-\epsilon)$ on both sides, we have 
\begin{align}
\epsilon^2-(c^2+2+\delta) \epsilon + c^2 \delta \geq 0,\label{eq: key2_eq}
\end{align}
which implies that $\epsilon \geq \frac{x + \sqrt{x^2-y}}{2}$ or  $\epsilon \in [0,\frac{x - \sqrt{x^2-y}}{2}]$, where $x=c^2+2+\delta$, and $y=4 c^2 \delta$.

For the case of $\epsilon \in [0,\frac{x - \sqrt{x^2-y}}{2}]$, since 
\begin{align}
(x - \sqrt{x^2-y})^2=2x^2-y-2x\sqrt{x^2-y} \geq 0,
\end{align}
we then obtain
\begin{align}
 \frac{y}{4x} \leq \frac{x - \sqrt{x^2-y}}{2}. \label{eq:  key3_eq}
\end{align}
Due to   $[0,\frac{y}{4x}] \subset  [0,\frac{x - \sqrt{x^2-y}}{2}]$ implied by \eqref{eq:  key3_eq}, we select
 $\epsilon=\frac{y}{4x}$ that satisfies \eqref{eq: key2_eq}. That is,
\begin{align}
\epsilon=\frac{y}{4x}=\frac{c^2 \delta}{c^2+2+\delta}
=\frac{c^2}{\frac{c^2}{\delta}+\frac{2}{\delta}+1}.\label{eq:  key4_eq}
\end{align}
Moreover, we have 
\begin{align}
c^2=(\mathbf a_{l_t}^T \mathbf v_{t-1})^2 \leq \| \mathbf a_{l_t}  \|_2^2  \| \mathbf v_{t-1} \|_2^2  \leq 4,\label{eq:  key5_eq}
\end{align}
where we have used the fact that $[\mathbf a_{l_t}]_i = 1$,  $[\mathbf a_{l_t}]_j = -1$ and $0$s elsewhere for the $l_t$th  edge in $\mathcal E_{\mathrm{selt}}$ that connects nodes $i$ and $j$.

Based on \eqref{eq:  key4_eq} and \eqref{eq:  key5_eq}, we have
\begin{align}
\epsilon=\frac{c^2}{\frac{c^2}{\delta}+\frac{2}{\delta}+1} \geq \frac{c^2}{\frac{6}{\delta}+1},
\end{align}
which  yields
\begin{align}
\lambda_{n-1}(\mathbf L_t)  \geq \lambda_{n-1}(\mathbf L_{t-1}) +\frac{c^2}{\frac{6}{\delta}+1}.
\end{align}
Recalling the definitions of $c$ and $\delta$, we finally obtain  \eqref{eq: inc_conn}. The proof is now complete.
\hfill $\blacksquare$

\section{Derivation of Inequality \eqref{eq: dif_Phi_new2}}\label{app: ineq}
Substituting \eqref{eq: z_it} into \eqref{eq: NET}, the term $\bar{\mathbf z}(t) - \mathbf z_i(t)$ becomes
\begin{align}
\bar{\mathbf z}(t) - \mathbf z_i(t) 
=& \sum_{s = 1}^{t-1} \sum_{j = 1}^n \left ( \frac{1}{n} - [ \boldsymbol \Phi(t-1,s)]_{ji} \right ) \mathbf g_j(s-1) \nonumber \\
&+ \sum_{j = 1}^n \frac{1}{n} \mathbf g_j(t-1) -\mathbf g_i(t-1).
\label{eq: z_zbar_it}
\end{align}
Since $f_i$ is $L$-Lipschitz continuous, we have
$\| \mathbf g_i(t) \|_* \leq L$ for all $i$ and $t$, and
\begin{align}
\| \bar{\mathbf z}(t) - \mathbf z_i(t) \|_* & \leq 
 L \sum_{s = 1}^{t-1}   \left \|   \boldsymbol \Phi(t-1,s) \mathbf e_i - \mathbf 1/n \right \|_1 + 2 L, \label{eq: dif_Phi}
\end{align}
where $\mathbf e_i$ is a basis vector with $1$ at the $i$th coordinate, and $0$s elsewhere.
For a  doubly stochastic matrix $\boldsymbol \Phi(t-1,s)$,   we have the following inequality \cite{ducaga12,matrix_bk}
\begin{align}
\|  \boldsymbol \Phi(t-1,s)  \mathbf e_i - \mathbf 1/n \|_1 \leq \sigma_2(\boldsymbol \Phi(t-1,s) ) \sqrt{n}. \label{eq: ineq_l1}
\end{align}
Based on \eqref{eq: dif_Phi} and \eqref{eq: ineq_l1}, we obtain  \eqref{eq: dif_Phi_new2}.  \hfill $\blacksquare$


\section{Proof of Lemma\,\ref{Lemma: sigma2}}
\label{app: sigma2}
Since $ \boldsymbol \Phi(t,s) $ is  doubly stochastic, 
we have
 $\sigma_1(\boldsymbol \Phi(t,s) ) = \lambda_1(\boldsymbol \Phi(t,s) ) = 1$ \cite[Ch.\,8]{matrix_bk}. 
The singular value decomposition of $ {\boldsymbol \Phi}(t,s)$ is given by
\[
\boldsymbol \Phi(t,s) = \mathbf U \boldsymbol \Gamma \mathbf V^T = \sum_{i=1}^n \sigma_i \mathbf u_i \mathbf v_i^T,
\]
where $\sigma_1 = 1$, and $\mathbf u_1 = \mathbf v_1 = \mathbf 1/\sqrt{n}$.

Consider a matrix deflation 
$
\tilde{ \boldsymbol \Phi}(t,s) = \boldsymbol \Phi(t,s) - \mathbf 1 \mathbf 1^T / n$,
we have
$
\sigma_1 ( \tilde{ \boldsymbol \Phi}(t,s) )  = \sigma_2( \boldsymbol \Phi(t,s) ) 
$.
Based on \cite[Theorem\,9]{Merikoski04}, we then obtain  
\begin{align}
\sigma_1 \left (    \tilde{\boldsymbol \Phi} (t,s+1) \tilde{\mathbf P}_s     \right ) \leq   \sigma_1( \tilde{\boldsymbol \Phi} (t,s+1) ) \sigma_1( \tilde{\mathbf P}_s)  , \label{eq: sigma1_Phits}
\end{align}
where $ \tilde{\mathbf P}_s  = \mathbf P_s - \mathbf 1 \mathbf 1^T / n $, and
\begin{align}
  \tilde{\boldsymbol \Phi} (t,s+1)    \tilde{\mathbf P}_s  = &[  \boldsymbol \Phi(t,s+1) - \mathbf 1 \mathbf 1^T / n ]  [{ \mathbf P}_s  - \mathbf 1 \mathbf 1^T / n] \nonumber \\
= & {  \boldsymbol \Phi(t,s+1) \mathbf P_s} - \mathbf 1 \mathbf 1^T /n = \boldsymbol \Phi(t,s) - \mathbf 1 \mathbf 1^T /n  \nonumber \\
= &  \tilde{ \boldsymbol \Phi}(t,s),~ \text{for all $s \leq t$.} \nonumber 
\end{align}
From  
 \eqref{eq: sigma1_Phits}, we obtain
$
\sigma_1 \left (  \tilde{ \boldsymbol \Phi}(t,s)  \right ) \leq  
\prod_{i = s}^t \sigma_1(\tilde{\mathbf P}_i )$,
which is equivalent to \eqref{eq: sigma2_Phi}.

\section{Proof of Proposition\,\ref{prop: net}}
\label{app: net_err}
Let $\delta \Def t - s +1$, from \eqref{eq: P_inc} and \eqref{eq: single_sigma2t} we  have 
\begin{align}
 \prod_{k={s}}^{t} \sigma_2( \mathbf P_k)  &\leq \prod_{k={0}}^{\delta-1} \sigma_2( \mathbf P_k) \nonumber \\
 &\leq  \prod_{k={0}}^{\delta-1}   \left [   
 \sigma_{2}(  \mathbf P_{0}  ) - \sum_{i = 1}^k u_i b_{i-1} (  \mathbf a_{l_i}^T \mathbf v_{i-1}  )^2
 \right ] ,  \label{eq: improvement_ex}
\end{align}
which yields
\begin{align}
 \prod_{k={0}}^{\delta-1 }  \frac{ \sigma_2( \mathbf P_k) }{\sigma_2(\mathbf P_0)}   
 \leq   \prod_{k={0}}^{\delta-1 }  \left ( 1 - \frac{\sum_{i = 1}^{k}  u_i b_{i-1} (\mathbf a_{l_i}^T \mathbf v_{i-1} )^2 }{\sigma_2(\mathbf P_0)} \right ).
\label{eq: ratio_beta}
\end{align}

Next, we introduce a variable $\beta$ defined as
\begin{align}
  \beta 
 \Def \prod_{k={0}}^{\delta-1 }  \left ( 1 - \frac{\sum_{i = 1}^{k}  u_i b_{i-1} (\mathbf a_{l_i}^T \mathbf v_{i-1} )^2 }{\sigma_2(\mathbf P_0)} \right )   
, \label{eq: beta_delta}
\end{align}
where by convention, $\beta = 1$ if $\delta = 1$.

Let $(\beta^*, \delta^*)$ be the solution of problem \eqref{eq: prob_delta_new}, we have
\begin{align}
& \delta^*  \geq 
\frac{\log{T\sqrt{n}}}{\log{ \sigma_2( \mathbf P_0)^{-1}}} -  \frac{\log{{\beta^*}^{-1}}}{\log{ \sigma_2( \mathbf P_0)^{-1}}}. 
\label{eq: delta_star_ex}
\end{align}
Based on \eqref{eq: improvement_ex} and  \eqref{eq: delta_star_ex}, for any $\delta \geq \delta^*$ we have
\begin{align}
 \prod_{i={s}}^{t} \sigma_2( \mathbf P_i)    \leq \prod_{i={0}}^{\delta^*-1} \sigma_2( \mathbf P_i)   \leq  \beta^* \sigma_2( \mathbf P_0)^{\delta^*} \leq \frac{1}{T\sqrt{n}}, \label{eq: beta_epsilon}
\end{align}
where  the last inequality is equivalent to  \eqref{eq: delta_star_ex}.

We split the right hand side of \eqref{eq: dif_Phi} at time $\delta^*$, and   obtain  
\begin{align}
 \| \bar{\mathbf z}(t) - \mathbf z_i(t) \|_*   \leq &
  L \sum_{s = 1}^{t -  1 -\delta^*}     \left \|   \boldsymbol \Phi(t-1,s) \mathbf e_i - \mathbf 1/n \right \|_1  \nonumber  \\
& +L\sum_{s = t - \delta^* }^{t-1}   \left \|   \boldsymbol \Phi(t-1,s) \mathbf e_i - \mathbf 1/n \right \|_1  + 2 L .
\label{eq: dif_Phi_new3_ex}
\end{align}
Since $ \left \|   \boldsymbol \Phi(t-1,s) \mathbf e_i - \mathbf 1/n \right \|_1 \leq 2$, the second term at the right hand side of \eqref{eq: dif_Phi_new3_ex} yields the following upper bound that is independent of $t$,
\begin{align}
L \sum_{s = t - \delta^* }^{t-1}   \left \|   \boldsymbol \Phi(t-1,s) \mathbf e_i - \mathbf 1/n \right \|_1 \leq 2 L \delta^* \label{eq: net1}
\end{align}
Moreover, based on \eqref{eq: ineq_l1} and \eqref{eq: sigma2_Phi},  the first term at the right hand side of \eqref{eq: dif_Phi_new3_ex} yields
\begin{align}
& L \sum_{s = 1}^{t -  1 -\delta^*}     \left \|   \boldsymbol \Phi(t-1,s) \mathbf e_i - \mathbf 1/n \right \|_1  \leq  L \sqrt{n} \sum_{s = 1}^{t -  1 -\delta^*}  \prod_{i={s}}^{t-1} \sigma_2( \mathbf P_i) \nonumber \\
& \leq L \sqrt{n} \sum_{s = 1}^{t -  1 -\delta^*}  \prod_{i={t - 1 - \delta^*}}^{t-1} \sigma_2( \mathbf P_i) 
\leq L \sqrt{n} \sum_{s = 1}^{t -  1 -\delta^*}  \prod_{i={0}}^{\delta^*-1} \sigma_2( \mathbf P_i) \nonumber \\
& \leq L \sqrt{n} \sum_{s = 1}^{t -  1 -\delta^*} \frac{1}{T\sqrt{n}},
\label{eq: net2}
\end{align}
where the last inequality holds due to \eqref{eq: beta_epsilon}, and 
 $\sum_{s = 1}^{t -  1 -\delta^*} \frac{1}{T} \leq 1$.

Substituting \eqref{eq: net1} and \eqref{eq: net2} into  \eqref{eq: delta_star_ex}, we obtain
\begin{align}  
 \| \bar{\mathbf z}(t) - \mathbf z_i(t) \|_*   & \leq 3L + 2L  \delta^*.
\label{eq: l2_error_ex}
\end{align}
Substituting \eqref{eq: l2_error_ex} into \eqref{eq: NET}, we obtain 
\eqref{eq: NET_tight_ex}. \hfill $\blacksquare$

\section{Proof of Corollary\,\ref{lemma: schedule}}\label{app: schedule}
Since $t_i  \leq \hat t_i  $ when $u_{t_i} = \hat u_{\hat t_i} = 1$, at a particular time $t$
we have
$
\sum_{k=1}^t u_{k} \geq \sum_{k=1}^t \hat u_{k}
$. Upon defining
\[
\beta \Def  \prod_{k={0}}^{\hat \delta^*-1 }  \left ( 1 - \frac{\sum_{j = 1}^{k} u_j b_{j-1} ( \mathbf a_{l_j}^T \mathbf v_{j-1} )^2}{\sigma_2(\mathbf P_0)} \right ),
\]
 we have
\[
\beta \leq  \prod_{k={0}}^{\hat \delta^*-1 }  \left ( 1 - \frac{\sum_{j= 1}^{k} \hat u_j b_{j-1} ( \mathbf a_{l_j}^T \mathbf v_{j-1} )^2}{\sigma_2(\mathbf P_0)} \right ) = \hat \beta^*.
\]
Therefore, 
$
 \hat \delta^* \geq \frac{\log{T\sqrt{n}}}{\log{ \sigma_2( \mathbf P_0)^{-1}}} -  \frac{\log{(\hat \beta^*)^{-1}}}{\log{ \sigma_2( \mathbf P_0)^{-1}}} \geq \frac{\log{T\sqrt{n}}}{\log{ \sigma_2( \mathbf P_0)^{-1}}} -  \frac{\log{(\beta)^{-1}}}{\log{ \sigma_2( \mathbf P_0)^{-1}}} 
$. This implies that the pair $( \beta, \hat \delta^*)$ is   a feasible point to problem \eqref{eq: prob_delta_new} under the schedule
$\{ u_t \}$. Therefore, $\delta^* \leq \hat \delta^*$. \hfill $\blacksquare$

\section{Proof of Theorem\,2}\label{app: rate_DDA}
Based on Theorem\,1, we obtain
\begin{align}
f( \hat{\mathbf x}_i(T)) - f(\mathbf x^*)\leq & \frac{1}{T \alpha_T} \psi(\mathbf x^*) + \frac{L^2}{2T} \sum_{t=1}^T \alpha_{t-1}   \nonumber \\
&+ \sum_{t=1}^T \frac{L^2\alpha_t}{T} \left ( 6 \delta^*
+  9   \right ).
\label{eq: Thr3_1}
\end{align}
Let  $\alpha_t = a/\sqrt{t}$ with convention $\alpha_0 = \alpha_1$ for some constant $a$. And applying $\sum_{t=1}^T t^{-1/2}\leq 2 \sqrt{T} $ and $ \psi(\mathbf x^*) \leq R^2$ into \eqref{eq: Thr3_1}, we obtain
\begin{align}
f( \hat{\mathbf x}_i(T)) - f(\mathbf x^*)\leq  \frac{ R^2}{a \sqrt{T}} 
+   \frac{a L^2 }{\sqrt{T}} \left ( 12 \delta^*
+  19   \right ). \label{eq: rule_c}
\end{align}
Substituting $a = {R \sqrt{1 - \sigma_2( \mathbf P_0)}}/{L}$ into \eqref{eq: rule_c}, we obtain  \eqref{eq: opt_net_edges}.
\hfill $\blacksquare$

\section{Proof of Proposition\,\ref{prop: conv_T_lb}}\label{app: time}
From \eqref{eq: cons_beta_delta_new}, we obtain
\begin{align}
\beta \geq  \left ( 1 - \frac{\sum_{i = 1}^{\ell} b_{i-1} ( \mathbf a_{l_i}^T \mathbf v_{i-1} )^2}{\sigma_2(\mathbf P_0)} \right )^{\delta},
\label{eq: beta_ineq_relax}
\end{align} 
where $\ell$ is the   number of scheduled edges. 
Substituting \eqref{eq: beta_ineq_relax} into \eqref{eq: beta_ineq_new}, we obtain 
\begin{align}
\delta   \geq   \frac{\log{T\sqrt{n}}}{\log{ \sigma_2( \mathbf P_0)^{-1}}} -  \frac{\delta\log{\alpha^{-1}}}{\log{ \sigma_2( \mathbf P_0)^{-1}}}, \label{eq: ineq_multip_delta}
\end{align}
which yields
\begin{align}
\delta^* \geq \frac{\log{T\sqrt{n}}}{\log{ \sigma_2( \mathbf P_0)^{-1}} + \log{\alpha^{-1}} }, \label{eq: delta_lb_K}
\end{align}
where  $\delta^*$ is the optimal solution of problem \eqref{eq: prob_delta_new},
$\alpha \Def 1 - \frac{\sum_{i = 1}^{\ell} b_{i-1} ( \mathbf a_{l_i}^T \mathbf v_{i-1} )^2}{\sigma_2(\mathbf P_0)}$, and $ \alpha \in [0,1]$ due to \eqref{eq: single_sigma2t}.


To find the convergence time for the desired $\epsilon$-accuracy, consider the following inequality suggested by Theorem\,2, 
\begin{align}
 \frac{RL \sqrt{1 - \sigma_2( \mathbf P_0)}}{\sqrt{T}}  
\delta^*  \leq \epsilon.  \label{eq: error_bd_T_thr}
\end{align}
Substituting \eqref{eq: delta_lb_K} into \eqref{eq: error_bd_T_thr}, we obtain a necessary condition to bound  the convergence time 
\begin{align}
  \frac{RL\sqrt{1 - \sigma_2( \mathbf P_0)}  }{\sqrt{T}} \frac{1}{\log{ \sigma_2( \mathbf P_0)^{-1}} + \log{\alpha^{-1}} }
\leq \epsilon. \label{eq: error_bd_T_2}
\end{align}
Since $
 \log{ ( \sigma_2( \mathbf P_0)^{-1}\alpha^{-1} )}   \geq 1 - \sigma_2( \mathbf P_0)\alpha
$, it is sufficient to consider
\begin{align}
\frac{RL \sqrt{1 - \sigma_2( \mathbf P_0)}}{\sqrt{T}}  
\frac{1}{ 1 - \sigma_2( \mathbf P_0)\alpha  }
  \leq \epsilon,
\end{align}
which yields
\begin{align}
T = \Omega \left ( \frac{1}{\epsilon^2}\frac{   1 - \sigma_2(\mathbf P_0)}{ ( 1 -\sigma_2( \mathbf P_0)\alpha )^2}
\right ). \label{eq: T_alpha_app}
\end{align}
The proof is now complete. \hfill $\blacksquare$

\end{document}